\documentclass[sigconf]{acmart}

\usepackage{amsthm}
\usepackage{amsmath}
\usepackage{subcaption}
\usepackage{paralist}
\usepackage{cases}
\usepackage{booktabs}
\usepackage{threeparttable}
\usepackage{epstopdf}
\usepackage{upgreek}
\usepackage{endnotes}
\usepackage{etoolbox}
\usepackage{algpseudocode}
\usepackage{xspace}
\usepackage{array}
\usepackage{enumitem}
\usepackage{balance}
\usepackage{multirow}
\usepackage{xcolor}
\usepackage{graphicx}
\usepackage{wrapfig}
\usepackage{enumitem}
\usepackage{tabularx}
\usepackage{float}
\usepackage{svg}
\usepackage{subcaption}
\usepackage{flushend}
\usepackage[ruled,linesnumbered]{algorithm2e}
\usepackage{stfloats}
\usepackage{afterpage}
\AtBeginDocument{%
  \providecommand\BibTeX{{%
    \normalfont B\kern-0.5em{\scshape i\kern-0.25em b}\kern-0.8em\TeX}}}

\setcopyright{acmlicensed}
\copyrightyear{2018}
\acmYear{2018}
\acmDOI{XXXXXXX.XXXXXXX}

\acmConference[Conference acronym 'XX]{Make sure to enter the correct
  conference title from your rights confirmation emai}{June 03--05,
  2018}{Woodstock, NY}
\acmISBN{978-1-4503-XXXX-X/18/06}

\begin{document}

\title{LLMLight: Large Language Models as Traffic Signal Control Agents}

\author{Siqi Lai}
\affiliation{\institution{The Hong Kong University of Science and Technology (Guangzhou)}}
\email{slai125@connect.hkust-gz.edu.cn}

\author{Zhao Xu}
\affiliation{\institution{The Hong Kong University of Science and Technology (Guangzhou)}}
\email{zxu674@connect.hkust-gz.edu.cn}

\author{Weijia Zhang}
\affiliation{\institution{The Hong Kong University of Science and Technology (Guangzhou)}}
\email{wzhang411@connect.hkust-gz.edu.cn}

\author{Hao Liu$^*$}
\affiliation{
\institution{The Hong Kong University of Science and Technology (Guangzhou)}
}
\email{liuh@ust.hk}

\author{Hui Xiong}
\affiliation{
\institution{The Hong Kong University of Science and Technology (Guangzhou)}
}
\email{xionghui@ust.hk}

\renewcommand{\shortauthors}{Siqi Lai, Zhao Xu, Weijia Zhang, Hao Liu, \& Hui Xiong}

\makeatletter
\def\@ACM@checkaffil{% Only warnings <<<<<<<<<<<<<<<<
    \if@ACM@instpresent\else
    \ClassWarningNoLine{\@classname}{No institution present for an affiliation}%
    \fi
    \if@ACM@citypresent\else
    \ClassWarningNoLine{\@classname}{No city present for an affiliation}%
    \fi
    \if@ACM@countrypresent\else
        \ClassWarningNoLine{\@classname}{No country present for an affiliation}%
    \fi
}
\makeatother

\newcommand{\fix}{\marginpar{FIX}}
\newcommand{\new}{\marginpar{NEW}}
\newcommand{\ie}{\emph{i.e.,}\xspace}
\newcommand{\eg}{\emph{e.g.,}\xspace}
\newcommand{\etc}{\emph{etc.}\xspace}
\newcommand{\etal}{\emph{et al.}\xspace}
\newcommand{\TODO}[1]{{\color{red}TODO: {#1}}}
\newcommand{\red}[1]{{\color{red}{#1}}}
\newcommand{\hao}[1]{{\color{blue}{#1}}}
\newcommand{\jia}[1]{{\color{purple}{#1}}}
\newcommand{\siqi}[1]{{\color[rgb]{0.62,0.54,0.22}{#1}}}
\newcommand\beftext[1]{{\color[rgb]{0.5,0.5,0.5}{BEFORE:#1}}}
\newcommand{\eat}[1]{}

\begin{abstract}
  Traffic Signal Control (TSC) is a crucial component in urban traffic management, aiming to optimize road network efficiency and reduce congestion. Traditional TSC methods, primarily based on transportation engineering and reinforcement learning (RL), often struggle with generalization abilities across varied traffic scenarios and lack interpretability. This paper presents LLMLight, a novel framework employing Large Language Models (LLMs) as decision-making agents for TSC. Specifically, the framework begins by instructing the LLM with a knowledgeable prompt detailing real-time traffic conditions. Leveraging the advanced generalization capabilities of LLMs, LLMLight engages a reasoning and decision-making process akin to human intuition for effective traffic control. Moreover, we build LightGPT\footnote{The model weights are available at \url{https://huggingface.co/lightgpt}.}, a specialized backbone LLM tailored for TSC tasks. By learning nuanced traffic patterns and control strategies, LightGPT enhances the LLMLight framework cost-effectively. Extensive experiments conducted on ten real-world and synthetic datasets, along with evaluations by fifteen human experts, demonstrate the exceptional effectiveness, generalization ability, and interpretability of LLMLight with LightGPT, outperforming nine baseline methods and ten advanced LLMs. Our project is available at \url{https://github.com/usail-hkust/LLMTSCS}.
\end{abstract}

% \thanks{$^*$ Corresponding author.}






\maketitle

%%%%%%%%% Introduction %%%%%%%%%
\section{Introduction}
Traffic congestion has emerged as a critical issue impacting both human society and the environment. As urban migration continues accelerating, city populations are expanding rapidly, further intensifying this challenge. This trend has necessitated a focused effort on optimizing Traffic Signal Control (TSC), a pivotal research area in intelligent transportation \cite{wu2023transformerlight, zhang2023irregular}. Efficient TSC not only promises substantial economic and environmental benefits but also enhances societal well-being. However, effectively managing traffic signals in complex and dynamic urban road networks remains daunting.

Existing work on TSC primarily falls into two categories: transportation \cite{koonce2008traffic, hunt1982scoot, lowrie1990scats, varaiya2013max, cools2013self} and Reinforcement Learning~(RL)-based approaches \cite{chen2020toward, wei2019colight, oroojlooy2020attendlight, chenguang2021prg, wu2021efficient, zhang2022expression, wei2019presslight, wu2023transformerlight}. Transportation methods primarily focus on crafting efficient heuristic algorithms, dynamically adapting traffic signal configurations based on lane-level traffic conditions. However, these methods heavily depend on manual design and often exhibit limited effectiveness in complex scenarios. The emergence of deep neural networks (DNNs) \cite{mnih2015human} led to the introduction of deep RL-based techniques to address this challenge \cite{abdulhai2003reinforcement, el2010agent}. These approaches have exhibited remarkable performance across various traffic scenarios. Nevertheless, RL-based methods also present several drawbacks. Primarily, they may struggle with limited generalization ability, particularly when transferring to larger road networks or under highly uncommon scenarios (\eg extreme high-traffic situations), as their training data only covers limited traffic situations. Additionally, RL-based methods lack interpretability due to the black-box nature of DNNs, which makes it hard to explain the rationale behind control actions under specific traffic conditions. This lack of transparency significantly hampers their feasibility for real-world deployment.

The recent emergence of Large Language Models (LLMs) has revolutionized several domains with their exceptional zero-shot learning and generalization capabilities, emulating human-like reasoning to solve complex tasks.
For instance, AutoGPT~\cite{autogpt} proposes to break tasks into multiple sub-goals and iterate until the main task is completed. In the transportation domain, GPT-Driver~\cite{mao2023gpt} proposes to instruct GPT-3.5 to tackle motion planning tasks in autonomous driving. For traffic control, PromptGAT~\cite{da2023llm} uses LLMs to generate human knowledge to help the DNN model understand long-tail cases (\eg extreme weather) in TSC tasks, aiming to bridge the gap between real-world and simulations.
Despite these advancements, existing research heavily relies on generalist LLMs~\cite{wang2024llm} (\eg ChatGPT) and primarily uses them as auxiliary tools~\cite{da2023llm, pang2024illm}. However, generalist LLMs often lack specialized domain knowledge in TSC, potentially limiting their effectiveness in professional applications. The direct application of LLMs as TSC agents for human-like decision-making remains largely unexplored. Addressing these issues involves overcoming two major challenges.
% Despite these advancements, existing research primarily utilizes LLMs as auxiliary tools~\cite{da2023llm, wang2024llm, pang2024illm} to enhance decision-making, the direct employment of LLMs as TSC agents for human-like decision-making remains largely unexplored. In particular, two major challenges arise toward this goal.

First, LLMs are typically pre-trained on large-scale natural language corpora and rarely incorporate non-textual traffic data, such as sensor readings and GPS trajectories \cite{llama2, chung2022scaling}. Despite their generalization capability across various tasks and domains, an inherent gap exists between real-time traffic data and linguistic understanding. The first challenge lies in enabling LLMs to comprehend real-time traffic dynamics and effectively interact with the traffic environment, which is critical for effective LLM-based TSC.

Second, selecting and developing an effective LLM for TSC poses another significant challenge. Generalist LLMs often lack specific domain knowledge and are prone to hallucination problems in professional fields \cite{huang2023hallucination, ji2023survey}. Although state-of-the-art LLMs such as GPT-4 demonstrate promising generalization abilities, their closed-source nature and substantial usage costs pose barriers to their optimization for real-time TSC tasks. Consequently, building a specialized LLM tailored for TSC tasks is crucial to deliver more effective and human-aligned control policies.

\begin{figure}[t]
\centering
\includegraphics[width=\columnwidth]{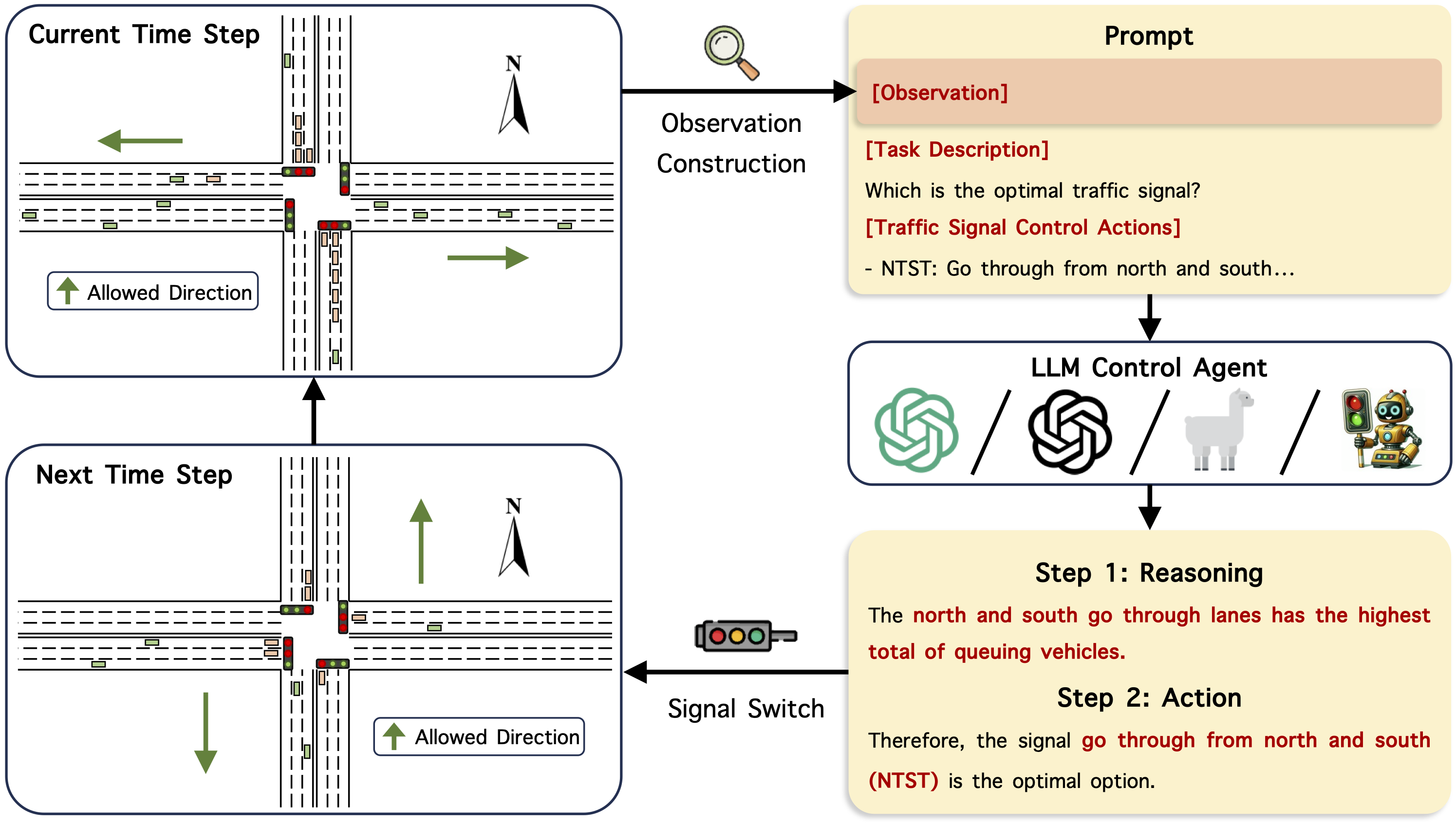}
\centering
\caption{The workflow of LLMLight.}
\label{fig:llmlight}
\vspace{-10pt}
\end{figure}

To this end, we introduce LLMLight, a traffic signal control agent framework based on LLMs, as depicted in Figure~\ref{fig:llmlight}. 
Specifically, we consider TSC as a partially observable Markov Game, where each agent, armed with an LLM, manages the traffic light at an intersection.
At each signal-switching time step, the agent collects traffic conditions of the target intersection and verbalizes them into human-readable text as real-time observation. 
Additionally, we incorporate task descriptions enriched with commonsense knowledge about a control strategy to aid the LLM's understanding of traffic management tasks. 
The combined real-time observation, task description, and control action space form a knowledgeable prompt that guides the LLM's decision-making. 
Then, the agent leverages Chain-of-Thought (CoT) reasoning to determine the optimal traffic signal configuration for the subsequent time step. 

Furthermore, we construct a specialized LLM, LightGPT, to enhance the LLMLight framework.
On the one hand, we propose imitation fine-tuning, enabling the specialized LLM to exploit high-quality control actions and underlying rationales derived from GPT-4.
On the other hand, we introduce a policy refinement process that utilizes a well-trained critic model to evaluate and improve the agent's actions.
The optimized LightGPT produces more effective control policies and showcases remarkable generalization ability across diverse traffic scenarios in a more cost-effective manner than the leading LLMs, including GPT-4.

We summarize the major contributions of this paper as follows:
(1) We pioneer the application of advanced LLMs as control agents, enabling human-like decision-making for interpretable traffic signal control. To our knowledge, this is the first exploration of directly using LLMs as decision-making agents in TSC.
(2) We introduce LLMLight, a tailored framework that engages various LLMs in effectively regulating traffic signals via knowledgeable instructions.
(3) We construct LightGPT, a specialized LLM backbone optimized for TSC tasks. This significantly enhances the framework's capability to derive effective control policies cost-effectively.
% To our knowledge, this is the first exploration of training LLMs as decision-making agents in TSC.
(4) Extensive experiments on ten traffic datasets, along with evaluations by fifteen human experts, demonstrate superior effectiveness, generalization capabilities, and interpretability of LLMLight and LightGPT compared to nine baseline methods and ten advanced generalist LLMs. Additionally, these experiments reveal key insights into the potential of integrating LLMs into intelligent transportation systems.

%%%%%%%%% Preliminaries %%%%%%%%%
\section{Preliminary}\label{sec:prelim}

% \begin{figure*}[t]
% \centering
% \includegraphics[width=\textwidth]{figs/prompt_archi.pdf}
% \centering
% \caption{The prompt template of LLMLight.}
% \label{fig:prompt_temp}
% \end{figure*}

In this section, we first introduce the key concepts in traffic signal control tasks.
\begin{definition}
    \textbf{Road network}: The road network is a directed graph consisting of intersection joints $I$ and lanes $L$. Lanes are categorized into go-through ($L_{go}$), left-turn ($L_{left}$), and right-turn ($L_{right}$) lanes, determined by the movement of vehicles on the lane. Each lane is divided into multiple segments $S=\left\{s_1,\dots,s_n\right\}$.
\end{definition}

\begin{definition}
    \textbf{Traffic signal control action}: The traffic signal control action corresponds to a specific signal phase, denoted as $a=set(L_{allow})$, where $L_{allow}$ is a group of allowed lanes without conflicting movements. $set(L_{allow})$ indicates the green light for activating the signal phase. At each signal-switching time step, the agent controlling the intersection chooses an action from $A=\left\{a_1,\dots,a_m\right\}$. The active signal phase allows vehicles in a specific lane group to pass, while other phases display the red light.
\end{definition}

%Figure \ref{fig:signal_phases} in the Appendix \ref{subsec:signal_phases} illustrates 
Please refer to Figure \ref{fig:signal_phases} (See Appendix \ref{subsec:signal_phases}) for more details of the most used setting of the intersection, lanes, and signal phases.

% \begin{figure}[t]
% \centering
% \includegraphics[width=\columnwidth]{figs/signal_phases.pdf}
% \centering
% \caption{An illustration of intersection, lanes, and signal phases. Lanes sharing a common color signify non-conflicting movements, and the passage is allowed in the corresponding signal phase. Signal phases include ETWT (go-through from east and west), ELWL (left-turn from east and west), NTST (go-through from north and south), and NLSL (left-turn from north and south).}
% \label{fig:signal_phases}
% \vspace{-10pt}
% \end{figure}

% Figure \ref{fig:signal_phases} illustrates the intersection, lanes, and signal phases utilized in our study.

% \begin{definition}
%     \textbf{Control action}: At each signal-switching time step, the agent controlling the intersection chooses an action from signal phase set $A=\left\{p_1,\dots,p_m\right\}$ to determine the next traffic signal configuration.
% \end{definition}

%%%%%%%%% Method %%%%%%%%%
\section{LLM for Traffic Signal Control}
This section first introduces the task formulation of traffic signal control in the LLM-empowered context. Then, we delve into the workflow of our proposed LLMLight, elucidating the methodology for prompting LLMs to perceive, analyze, and regulate traffic at the target intersection directly. Finally, we elaborate on the training procedures implemented for developing LightGPT, aiming to augment the backbone LLM's domain expertise in traffic management, thereby generating proficient control policies adaptable to diverse traffic scenarios.

\subsection{Task Formulation}
We define traffic signal control as a partially observable Markov Game \cite{wei2019colight, wei2019presslight}. The LLM-based agent manages the traffic light of an intersection with policy $\pi_\theta$. Based on the observation $o$ of the real-time traffic condition at the intersection, traffic scenario description $d_{scene}$, task descriptions $d_{task}$, commonsense knowledge $d_{know}$, and control action space $A$, the LLM outputs the reasoning trajectory $Y$ with the control action $a$ that regulates the traffic signal of the intersection, with the long-term goals of maximizing the transportation efficiency within the road network:
\begin{align}
    (Y, a) = \pi_{\theta}(Prompt(o, d_{scene}, d_{task}, d_{know}, A)).
\end{align}

\subsection{Workflow of LLMLight}
This subsection details the workflow of our proposed LLMLight, including (1) \textsl{observation collection}: it collects the traffic condition of the intersection; (2) \textsl{commonsense augmented prompt construction}: it composes a commonsense knowledge enhanced prompt, instructing the LLM to devise the optimal traffic signal configuration; (3)~\textsl{reasoning and action execution}: the LLM engages a decision-making process by employing the constructed prompt, and subsequently executes its selected control action.

\subsubsection{Observation Feature Construction}
We collect traffic condition features that can be easily obtained from the real-world traffic control environment:

\begin{itemize}[leftmargin=0.5cm]
    \item \textbf{Queuing vehicle count}: Vehicles with speeds slower than the threshold $v_{stop}$ are considered as queuing vehicles. We count their number in lane $l$ as $n^l_q$.
    \item \textbf{Approaching vehicle count}: Vehicles faster than $v_{stop}$ are considered as approaching vehicles. We count the number of approaching vehicles in the segment $s$ of lane $l$ as $n^l_s$.
\end{itemize}
Subsequently, these features on every incoming lane (where the vehicle enters the intersection) are verbalized into human-readable text as the observation $o$ of the LLM.

\begin{figure}[t]
  \begin{subfigure}{\columnwidth}
    \centering
    \includegraphics[width=\linewidth]{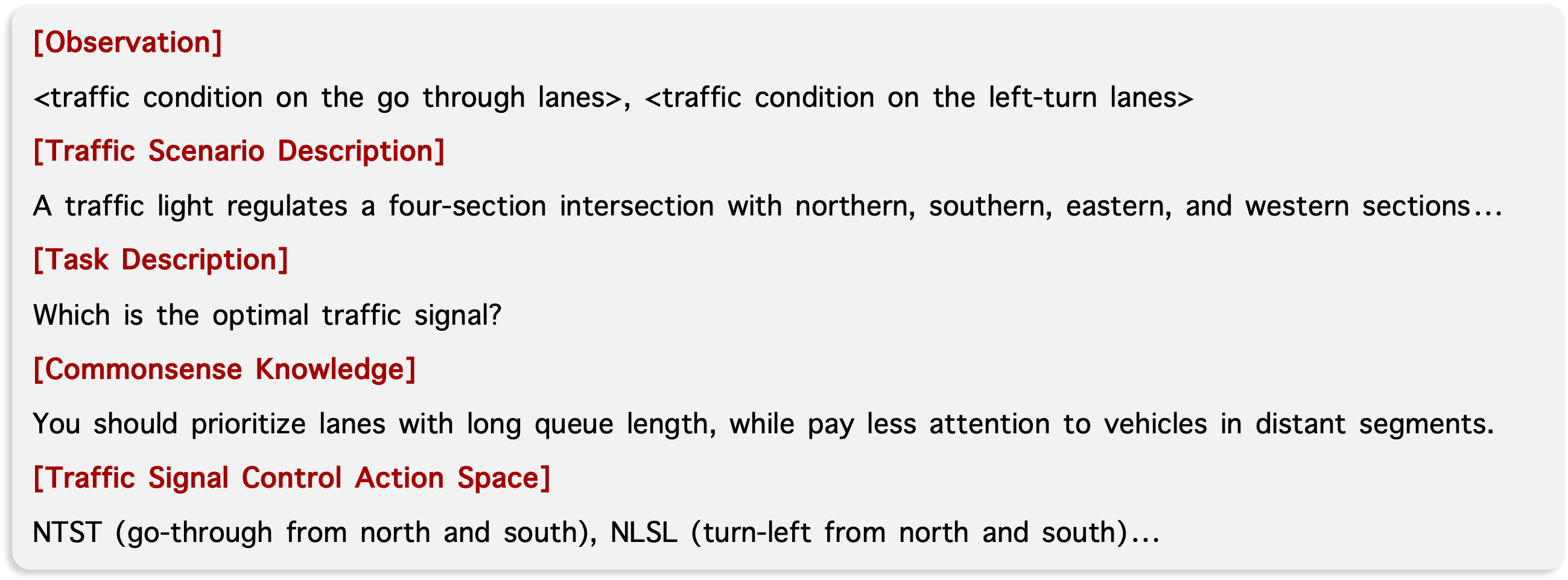}
    \caption{The prompt template.}
    \label{fig:prompt_achi}
  \end{subfigure}
  \begin{subfigure}{\columnwidth}
    \centering
    \includegraphics[width=\linewidth]{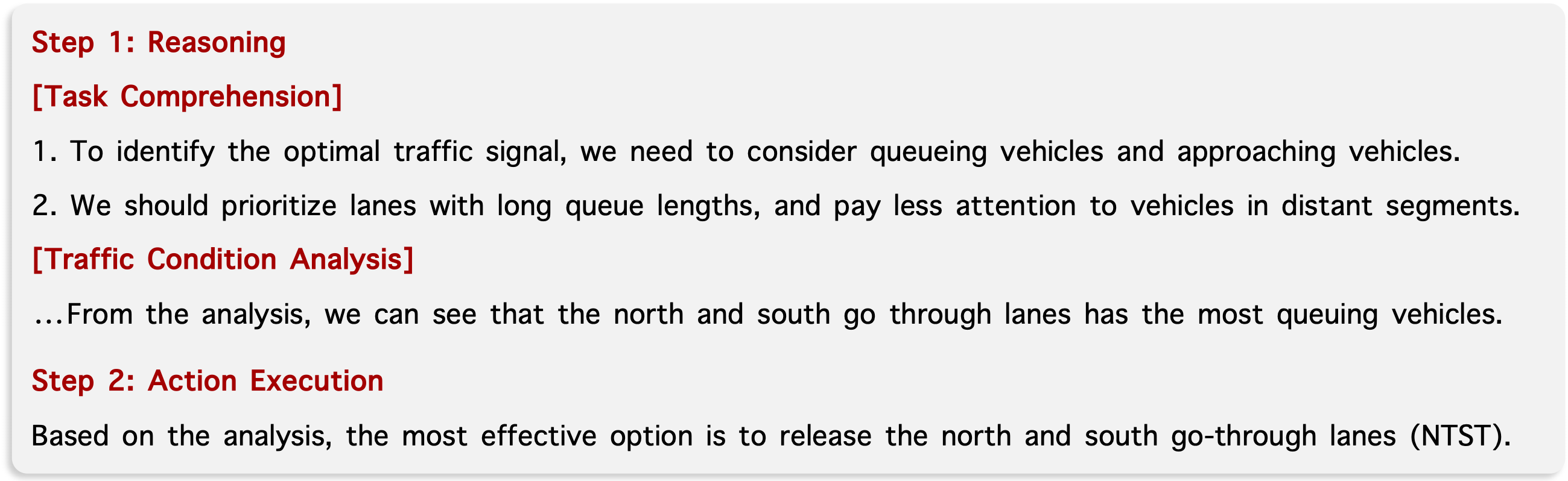}
    \caption{The reasoning and action execution process.}
    \label{fig:reasoning_archi}
    \vspace{-5pt}
  \end{subfigure}
  \caption{The prompt template of LLMLight and decision-making process of the backbone LLM. In this scenario, the intersection connects four lane sections, including north, south, east, and west.}
  \label{fig:prompt_reasoning}
\vspace{-10pt}
\end{figure}

\subsubsection{Commonsense Augmented Prompt Construction}\label{subsubsec:prompt_gen}
In addition to observation, we provide the LLM with comprehensive information essential in addressing traffic signal control, including traffic scenario description $d_{scene}$, task descriptions $d_{task}$, and control action space $A$. This enables the LLM to thoroughly understand the task, facilitating reasonable control actions. Furthermore, we integrate commonsense knowledge $d_{know}$ to mitigate generalist LLMs' limitations in traffic management, a domain they are not explicitly trained for. Specifically, this knowledge dictates that the agent needs to prioritize lanes with extended queue lengths while paying less attention to vehicles distant from the intersection. Formally, we denote the composed prompt as $X=Prompt(o, d_{scene}, d_{task}, d_{know}, A)$. A glimpse of the prompt template is shown in Figure \ref{fig:prompt_achi}, and a detailed example is provided in Table \ref{tab:prompt} (see Appendix \ref{subsec:prompt}).

\subsubsection{Reasoning and Action Execution}
We instruct the LLM to perform zero-shot reasoning using the constructed prompt. Its decision-making unfolds in two key steps: reasoning and action execution. Initially, it embarks on comprehensive analysis, assimilating the given task, leveraging commonsense knowledge, and assessing the current traffic conditions across various lanes. Subsequently, it selects the signal for the lanes experiencing the highest congestion, thereby optimizing traffic flow and ensuring the smooth passage of vehicles. In this manner, LLMLight can not only make effective control actions but also distinguishes itself by providing detailed rationales for each decision, contributing to a more explainable and transparent traffic control system. Formally, we denote the reasoning and action execution as $(Y, a) = \pi_\theta(X)$. An example of the backbone LLM's decision-making process is shown in Figure \ref{fig:reasoning_archi}. We conduct an interpretability analysis in Subsection \ref{subsec:interpret}.

\subsection{Backbone LLM Training}
\begin{figure}[t]
\centering
\includegraphics[width=0.48\textwidth]{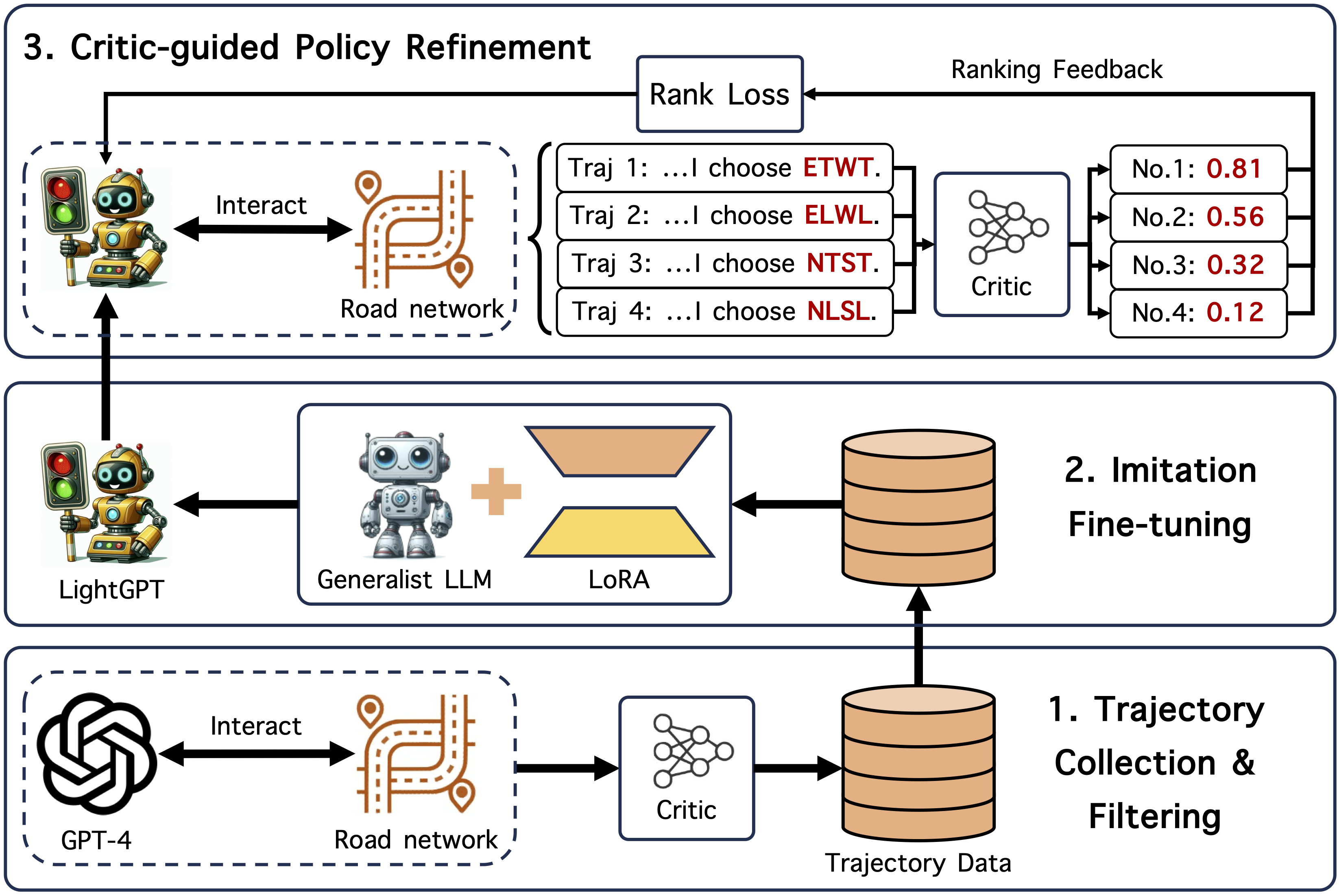}
\centering
\caption{The training procedure of LightGPT.}
\label{fig:training}
\vspace{-10pt}
\end{figure}

Fine-tuning has become a common routine to align LLM with diverse downstream applications \cite{chen2023fireact, zeng2023agenttuning} by augmenting the domain expertise in the specific realm. 
% While the state-of-the-art LLM, GPT-4, offers a feasible choice as the traffic signal control agent, it is closed-sourced and presents challenges in fine-tuning the generated control policy. Additionally, its cost is comparatively higher than locally deployable language models. Therefore,
We propose a training procedure to optimize an LLM specifically tailored for traffic signal control, leading to the development of LightGPT. Figure \ref{fig:training} presents the training workflow. It primarily consists of three stages: (1) \textsl{reasoning trajectory collection and filtering}: it collects the CoT reasoning trajectory of GPT-4 for imitation fine-tuning, then we select the trajectories that most align with the long-term goal to ensure the data quality; (2) \textsl{imitation fine-tuning}: it utilizes the filtered reasoning trajectories and the corresponding actions taken by GPT-4 to train the backbone LLM; (3) \textsl{critic-guided policy refinement}: it further improves the decision-making process of the LLM by fine-tuning with the feedback of a well-trained critic model.

\subsubsection{Trajectory Collection and Filtering}\label{subsubsec:trajectory_collection}
Following the prompt construction process detailed in Section \ref{subsubsec:prompt_gen}, we first instruct GPT-4 to interact with simulated traffic environments for gathering reasoning trajectories for subsequent imitation fine-tuning. To ensure the quality of the collected data, we employ a filtering strategy that identifies reasoning trajectories aligning most closely with the long-term objective of traffic signal control, such as minimizing future queue length. This filter operation is achieved by aligning with a pre-trained action-value network (the architecture is derived from \cite{zhang2022expression}), which is trained by optimizing Bellman equation within the simulation environment:
\begin{align}
    \mathcal{L}_{value} = \mathbb{E}[(R(o_t, a_t) + \gamma \max \limits_{a_{t+1}} Q(o_{t+1}, a_{t+1}) - Q(o_t, a_t))^2],
\end{align}
where $o_t$ and $a_t$ are the observation and action chosen at the signal-switching time step $t$, and $\gamma \in [0,1]$ is the discount factor. $R(o, a)$ is the reward function that provides feedback for executing action $a$ under observation $o$ (\eg the negative of queue length). $Q(o, a)$ is the action-value function estimating the cumulative future reward earned after executing $a$. Actions estimated to yield higher future rewards suggest greater alignment with the long-term goal.

Subsequently, the trained action-value function is utilized as a critic to evaluate the decision-making of GPT-4. We selectively retain only reasoning trajectories where the selected actions are associated with the highest estimated future rewards, as:
\begin{align}
    \big\{(X_t, Y_t, a_t)|a_t = \mathop{argmax} \limits_{a\in A} Q(o_t, a)\big\}^T_{t=0},
\end{align}
where $T$ is the simulation duration, $X_t$ is prompt, and $Y_t$ is the reasoning trajectory of GPT-4.

\subsubsection{Imitation Fine-tuning}
In the initial stage of backbone LLM training, we employ an imitation learning process, allowing the LLM to train on the control actions taken by GPT-4 and their underlying rationales. We treat prompt $X$ as fine-tuning instructions and GPT-4's reasoning trajectory $Y=[Y;a]$ containing its selected action $a$ as the anticipated response. Negative log-likelihood (NLL) is adopted as the loss function:
\begin{align}
    \mathcal{L}_{IFT} = -\sum_w^{|Y|} log P_{\pi_\theta}(y_w | X, Y_{<w}),
\end{align}
where $P_{\pi_\theta}(y_w | X, Y_{<w})$ is the probability of being prompted with $X$ to generate the token $y_w$. Due to the enormous GPU memory usage of full model training, we adopt parameter-efficient fine-tuning (PEFT) techniques, specifically low-rank adaptation (LoRA) \cite{hu2021lora}, in our training. Instead of adjusting all model weights, LoRA introduces low-rank matrices to model the weight updates, significantly reducing the number of parameters to be trained. This approach not only decreases computational and storage costs but also preserves the original model's performance.

\subsubsection{Critic-guided Policy Refinement}
To further enhance the effectiveness of the LLM's control policy, we propose a policy refinement procedure, which adjusts the LLM's reasoning trajectory for directing better actions. Similarly, we use the pre-trained action-value function elaborated in Section \ref{subsubsec:trajectory_collection} as the critic model to evaluate the actions derived by the backbone LLM. Higher evaluation scores correspond to greater estimated future rewards. Subsequently, we implement an alignment fine-tuning algorithm to adjust the reasoning trajectory, ultimately guiding the LLM towards actions that yield higher scores.
Although previous studies \cite{ouyang2022training, carta2023grounding} prove RL (\eg PPO \cite{schulman2017proximal}) is a potent tool for alignment, this method proved challenging to converge and yielded unsatisfactory results, as validated by \cite{wang2023making} and the experiment in Appendix \ref{subsec:ppo_exp}. Alternatively, we leverage a ranking loss \cite{wang2023making, yuan2023rrhf} function to achieve this goal.

Specifically, we have $k$ different reasoning trajectories $\mathbf{Y}=\left\{Y_i\right\}^k_{i=1}$ sampled by policy $\pi_\theta$ under instruction $X$. The critic gives scores for actions derived by each trajectory $Y_i$ as $q_i = Q(o, a_i)$. Afterwards, the token-averaged log-likelihood value of each $Y_i$ is derived to denote the corresponding probability for $\pi_\theta$ to generate $Y_i$:
\begin{align}
    p_i = \frac{1}{{|Y_i|}} \sum^{|Y_i|}_w logP_{\pi_\theta}(y_{i,w}|X,Y_{i,<w}).
\end{align}
Following \cite{wang2023making}, we adopt ranking-feedback loss with a boundary-constrained term (RBC) for generating reasoning trajectories that result in actions with higher scores:
\begin{align}
    \mathcal{L}_{CGPR} = log\big\{ 1 + \sum_{q_i > q_j} \big[ e^{(p_j - p_i)} + e^{(2p_{j*} - 2\beta - p_i - p_j)}\big] \big\},
\end{align}
where $p_{j*} = \min \limits_{q_k > q_j} p_k$ is the probability of the least favorable reasoning trajectory rated higher than $Y_j$, $\beta$ is a hyper-parameter, $e^{(p_j - p_i)}$ is the alignment term to elevate the probability of generating trajectories with actions rated higher scores, and $e^{(2p_{j*} - 2\beta - p_i - p_j)}$ is the constraint term for prohibiting performance degradation.

% It is crucial to be aware that while this refinement procedure is guided by a pre-trained critic model developed from RL, this doesn't imply that the backbone LLM learns exclusively from it. The alignment process is trained on reasoning trajectories sampled by the LLM's policy. The objective is to encourage the model to discern which trajectory is more reasonable and leads to decisions more closely aligned with long-term goals. However, it's important to note that we do not provide any ground-truth labels, thus the LLM is not bound to follow the critic's predicted ``best'' action directly. The experiments presented in Subsection \ref{subsec:generalization} validate that RL models experience performance degradation under varied traffic scenarios, whereas LLMLight equipped with LightGPT remains robust based on our proposed training procedure.

It is crucial to be aware that while this refinement process is guided by a pre-trained critic model developed from RL, it does not imply that the backbone LLM learns exclusively from it. The alignment process is trained on reasoning trajectories sampled by the LLM's policy, aiming to help the model identify trajectories with reasonable decisions that align closely with long-term goals. However, it is important to note that we do not provide any ground-truth labels in the training, thus the LLM is not bound to follow the critic's predicted ``best'' action directly. Experiments in Subsection \ref{subsec:generalization} demonstrate that RL models show performance degradation in varied traffic scenarios. In contrast, LLMLight equipped with LightGPT remains robust based on our proposed training procedure.

%%%%%%%%% Experiments %%%%%%%%%
\section{Experiments}\label{sec:experiment}
In this section, we conduct extensive experiments to evaluate our proposed LLMLight by answering the following research questions:
\begin{itemize}[leftmargin=0.5cm]
    \item \textbf{RQ1}: How is the \textbf{effectiveness} of LLMLight compared with traditional transportation and RL methods?
    
    \item \textbf{RQ2}: How is the \textbf{generalization} ability of LLMLight across different traffic scenarios and vehicle volumes?
    
    \item \textbf{RQ3}: How is the \textbf{interpretability} of LLMLight in providing rationales behind the selected control actions?
\end{itemize}

% \vspace{-5pt}
\subsection{Experimental Settings}

\subsubsection{Datasets}
Our experiments use seven real-world \cite{wei2019survey} and three synthetic traffic flow datasets. Statistics for these datasets are detailed in Table \ref{tab:dataStats}. Visualizations of road networks are presented in Figure \ref{fig:roadnet} (see Appendix \ref{subsec:road_network}).

\begin{itemize}[leftmargin=0.5cm]
    \item \textbf{Jinan}: This dataset is from Dongfeng sub-district, Jinan, China, featuring 12 intersections. It includes three traffic flow datasets recorded over three different periods. The area covered is 400 meters east-west and 800 meters north-south.
    
    \item \textbf{Hangzhou}: Collected from Gudang sub-district, Hangzhou, China, this dataset features 16 intersections and includes two traffic flow datasets from different periods. The area spans 800 meters east-west and 600 meters north-south.
    
    \item \textbf{New York}: Collected in the Upper East Side of Manhattan, New York, USA, using taxi trip data, this dataset covers 196 intersections with lanes spanning 300 meters. It includes two large-scale traffic flow datasets from different periods to evaluate the scalability of different methods.
    
    \item \textbf{Extreme High-traffic}: Synthesized based on real-world traffic in Jinan and Hangzhou, these datasets feature approximately four times more vehicles per five minutes than the original datasets to simulate high-traffic scenarios.

    \item \textbf{24-hour Cycle}: Modeled on Jinan's actual traffic flows, this dataset replicates realistic daily traffic fluctuations over a 24-hour period, including rush hours \cite{Adeyemi2021ExploringTR} from 6:00 am to 10:00 am and 3:00 pm to 8:00 pm.
    
    % To accurately replicate real-world daily traffic patterns over a 24-hour period, we generate a dataset modeled on actual traffic flows in Jinan. This dataset provides realistic fluctuations of daily traffic, including distinct morning and evening traffic rush hours. Following , we define rush hours as 6:00 am to 10:00 am and 3:00 pm to 8:00 pm.
\end{itemize}

\begin{small}
\begin{table}[t]
\renewcommand{\arraystretch}{1.1}
 \small
 \centering
 \caption{Statistics of datasets.}
 \vspace{-5pt}
 \resizebox{0.48\textwidth}{!}{
  \begin{tabular}{c|c|c|cccc}
   \toprule
   \multirow{2}{*}{Flow dataset} & \# of inter- & \multirow{2}{*}{\# of vehicles} & \multicolumn{4}{c}{Arrival rate (vehicles/5min)} \\
   \cline{4-7}
   & sections & & Mean & Std & Max & Min \\
   \midrule
   % 添加表格内容
   Jinan 1 & \multirow{5}{*}{12} & 6295 & 523.67 & 98.52 & 671 & 255 \\
   Jinan 2 & & 4365 & 362.83 & 74.81 & 493 & 236 \\
   Jinan 3 & & 5494 & 456.92 & 46.38 & 544 & 362 \\
   Jinan Extreme & & 24000 & 1999.08 & 44.14 & 2097 & 1932 \\
   Jinan 24-hour & & 58200 & 201.09 & 160.88 & 569 & 2 \\
   \midrule
   Hangzhou 1 & \multirow{3}{*}{16} & 2983 & 247.67 & 40.44 & 332 & 211\\
   Hangzhou 2 & & 6984 & 581.08 & 318.43 & 1145 & 202 \\
   Hangzhou Extreme & & 24000 & 1999.08 & 40.66 & 2073 & 1934 \\
   \midrule
   New York 1 & \multirow{2}{*}{196} & 11058 & 849.69  & 174.06 & 964  & 382 \\
   New York 2 & & 16337 & 1255.77 & 264.85 & 1440 & 475 \\
   \bottomrule
  \end{tabular}}
  \label{tab:dataStats}
  % \vspace{-10pt}
\end{table}
\end{small}

% \vspace{-5pt}
\subsubsection{Data Anonymization}
The datasets we used do not include any identifiable driver information, such as names or contact details. These datasets are open-sourced and publicly available at \cite{traffic-signal-control}.

% \vspace{-5pt}
\subsubsection{Environment Settings}
% We conduct experiments on CityFlow~\cite{zhang2019cityflow}, an open-source simulator widely recognized for benchmarking traffic signal control policies. CityFlow supports large-scale, city-wide traffic simulations and features an interactive render for monitoring. It provides extensive APIs to access traffic state features and execute control actions. In our experiments, each intersection connects four lane sections: East, West, North, and South. We utilize four signal phases as corresponding control actions, including ETWT (go-through from east and west), ELWL (left-turn from east and west), NTST (go-through from north and south), and NLSL (left-turn from north and south). The turn-right movements are always allowed in simulation. The green signal phase duration is set to thirty seconds. Following existing studies~\cite{zhang2022expression, wei2019colight}, a three-second yellow signal and a two-second all-red time follow each green signal to prepare the transition. The simulation spans twenty-four hours for the 24-hour cycle dataset, while others span one hour.

We conduct experiments on CityFlow, an open-source simulator widely used for benchmarking traffic signal control policies. CityFlow supports large-scale, city-wide traffic simulations and features an interactive render for monitoring. It provides extensive APIs to access traffic state features and execute control actions. In our experiments, each intersection connects four lane sections: East, West, North, and South. We utilize four signal phases as corresponding control actions: ETWT (go-through from east and west), ELWL (left-turn from east and west), NTST (go-through from north and south), and NLSL (left-turn from north and south). Right-turn movements are always allowed in the simulation. The green signal phase duration is set to thirty seconds. Following existing studies~\cite{zhang2022expression, wei2019colight}, a three-second yellow signal and a two-second all-red time follow each green signal to prepare the transition. The simulation spans twenty-four hours for the 24-hour cycle dataset, while others span one hour.

% \vspace{-5pt}
\subsection{Metrics}
Following previous studies \citep{zhang2022expression}, we leverage \textsl{average travel time} (ATT), \textsl{average queue length} (AQL), and the \textsl{average waiting time} (AWT) of vehicles to evaluate the performance of different policies made by traffic signal control agents.
\begin{itemize}[leftmargin=0.5cm]
    \item \textbf{Average traveling time (ATT)} quantifies the average duration of all the vehicles traveling from their origins to their respective destinations.
    
    \item \textbf{Average queue length (AQL)} is defined as the average number of queuing vehicles waiting in the road network.
    
    \item \textbf{Average waiting time (AWT)} quantifies the average queuing time of vehicles at every intersection in the road network.
\end{itemize}

% \vspace{-5pt}
\subsection{Compared Models}
For transportation methods, we adopt Random, FixedTime, and Maxpressure as baselines. For RL methods, we compare MPLight, AttendLight, PressLight, CoLight, Efficient-CoLight, and Advanced-CoLight with our proposed method. For LLMs, we utilize GPT-4, ChatGPT-3.5, Qwen, Llama2, and Llama3 as backbone models for LLMLight. For open-source LLMs, such as Qwen and Llama, we evaluate their performance both with and without fine-tuning. Due to the high training and deployment costs of models larger than 70B parameters (\eg Llama2-70B or Qwen2-72B) in real-world traffic scenarios, we only test and report their performances without fine-tuning. Please refer to Appendix \ref{subsec:baseline} for more details about the abovementioned models.

% \vspace{-5pt}
\subsection{Model Settings}
All RL methods are trained with uniform hyperparameters, encompassing a learning rate of $1 \times 10^{-3}$, a replay buffer size of $12000$, a sample size of $3000$, and a hidden size of $20$. The \texttt{top\_p} value is standardized at $1.0$ for all LLMs. To stabilize the LLMs' output, a temperature setting of zero is implemented for ChatGPT-3.5 and GPT-4, while open-sourced LLMs are set to $0.1$. The settings of LoRA, including a rank of 8 and a scaling factor $\alpha$ of 16, were set across all training stages. In the imitation fine-tuning, the batch and mini-batch sizes are set to $128$ and $16$, respectively. We sample four reasoning trajectories under each instruction for critic-guided policy refinement. The fine-tuning learning rate is set to $3\times 10^{-4}$.
% The hyperparameters of low-rank adaptation (LoRA) are set uniformly in every fine-tuning stage, encompassing the rank of 8 and the $\alpha$ coefficient of 16.

\begin{table*}[!]
\centering
\caption{Overall performances of LLMLight and previous traditional methods on Jinan and Hangzhou datasets. The best, second-best, and third-best results are highlighted through \textbf{boldface}, \underline{\underline{double underline}}, and \underline{underline}, respectively.}
% \vspace{-5pt}
\label{tab:overall}
\setlength{\tabcolsep}{4pt}
\resizebox{\textwidth}{!}{
\begin{tabular}{c|ccc|ccc|ccc|ccc|ccc}
\toprule
\multirow{4}{*}{Models}&
\multicolumn{9}{c}{\textbf{Jinan}} &\multicolumn{6}{c}{\textbf{Hangzhou}}\cr
\cmidrule(lr){2-10} \cmidrule(lr){11-16}
& \multicolumn{3}{c}{1} & \multicolumn{3}{c}{2} & \multicolumn{3}{c}{3} & \multicolumn{3}{c}{1} & \multicolumn{3}{c}{2}\cr
\cmidrule(lr){2-4} \cmidrule(lr){5-7} \cmidrule(lr){8-10} \cmidrule(lr){11-13} \cmidrule(lr){14-16}
& ATT & AQL & AWT & ATT & AQL & AWT & ATT & AQL & AWT & ATT & AQL & AWT & ATT & AQL & AWT \\
\midrule
\multicolumn{16}{c}{\textbf{Transportation Methods}} \\
\midrule
Random                   & 597.62 & 687.35 & 99.46 & 555.23 & 428.38 & 100.40 & 552.74 & 529.63 & 99.33  & 621.14 & 295.81 & 96.06 & 504.28 & 432.92 & 92.61 \\
FixedTime                & 481.79 & 491.03 & 70.99 & 441.19 & 294.14 & 66.72  & 450.11 & 394.34 & 69.19  & 616.02 & 301.33 & 73.99 & 486.72 & 425.15 & 72.80 \\
Maxpressure              & 281.58 & 170.71 & \underline{44.53} & 273.20 & 106.58 & \textbf{38.25}  & 265.75 & 133.90 & \underline{\underline{40.20}}  & 325.33 & 68.99  & 49.60 & 347.74 & 215.53 & 70.58 \\
\midrule
\multicolumn{16}{c}{\textbf{RL Methods}} \\
\midrule
MPLight                  & 307.82 & 215.93 & 97.88 & 304.51 & 142.25 & 90.91  & 291.79 & 171.70 & 89.93  & 345.60 & 84.70  & 81.97 & 358.56 & 237.17 & 100.16 \\
AttendLight              & 291.29 & 186.25 & 61.73 & 280.94 & 115.52 & 52.46  & 273.02 & 144.05 & 55.93  & 322.94 & 66.96  & 55.19 & 358.81 & 239.05 & 72.88 \\
PressLight               & 291.57 & 185.46 & 50.53 & 281.46 & 115.99 & 47.27  & 275.85 & 148.18 & 54.81  & 364.13 & 98.67  & 90.33 & 417.01 & 349.25 & 150.46 \\
CoLight                  & 279.60 & 168.53 & 58.87 & 274.77 & 108.28 & 54.14  & 266.39 & 135.08 & 53.33  & 322.85 & 66.94  & 61.82 & 342.90 & 212.09 & 99.74 \\
Efficient-CoLight        & 277.11 & 163.60 & \underline{\underline{43.41}} & \underline{269.24} & \underline{102.98} & \underline{\underline{39.74}}  & 262.25 & 129.72 & \textbf{39.99}  & 311.96 & \underline{58.20}  & \textbf{36.83} & \underline{333.27} & \underline{189.65} & \textbf{61.70} \\
Advanced-CoLight         & \underline{\underline{274.67}} & \underline{\underline{160.85}} & 49.30 & \underline{\underline{268.25}} & \underline{\underline{102.12}} & 41.11  & \underline{260.66} & \underline{127.83} & 43.54  & \textbf{304.47} & \textbf{52.94}  & \underline{41.75} & \textbf{329.16} & \textbf{186.34} & 76.59 \\
\midrule
\multicolumn{16}{c}{\textbf{LLMLight (with Generalist LLMs)}} \\
\midrule
% Llama2-13B             & 508.72 & 567.61 & 169.67 & 435.17 & 290.61 & 156.30 & 452.24 & 416.38 & 240.99 & 427.37 & 152.18  & 157.47 & 432.21 & 388.59 & 220.67 \\
Qwen2-72B              & 291.95 & 185.20 & 58.65  & 277.78 & 112.13 & 53.12  & 274.33 & 144.95 & 54.94  & 321.91 & 65.75  & 66.52  & 342.37 & 206.07 & 100.55 \\
Llama2-70B             & 353.03 & 286.52 & 85.96  & 324.52 & 162.34 & 99.87  & 320.41 & 210.13 & 100.90 & 357.95 & 93.21  & 106.63 & 361.53 & 250.69 & 121.58 \\
Llama3-70B             & 290.19 & 186.43 & 61.09  & 277.49 & 112.86 & 51.76  & 271.60 & 142.95 & 54.55  & 325.85 & 69.42  & 71.51  & 339.23 & 198.82 & 78.81 \\
ChatGPT-3.5            & 536.79 & 952.93 & 155.62 & 524.81 & 393.21 & 179.34 & 501.36 & 479.50 & 154.19 & 463.04 & 181.95 & 191.87 & 418.75 & 336.47 & 130.56 \\
GPT-4                  & 275.26 & 160.93 & 46.61 & 271.34 & 105.22 & 47.55 & 264.70 & 132.53 & 46.16 & 318.71 & 62.84 & 58.09 & 335.81 & 193.32 & 66.02 \\
\midrule
\multicolumn{16}{c}{\textbf{LLMLight (with LightGPTs)}} \\
\midrule
% LightGPT (IFT)      & 277.60 & 166.17 & 48.21 & 272.24 & 104.93 & 48.78 & \underline{261.34} & \underline{129.27} & 45.02 & 316.4 & 61.55 & 49.92 & 333.42 & 190.43 & 68.99 \\
LightGPT (Qwen2-0.5B)    & 296.71 & 193.93 & 46.80 & 292.99 & 129.93 & 80.63 & 290.08 & 169.56 & 91.20 & 328.11 & 70.90 & 84.49 & 351.21 & 233.16 & 102.39 \\
LightGPT (Qwen2-7B)      & 275.92 & 163.34 & 48.56  & 270.41 & 104.40 & 44.93 & 263.10 & 130.94 & 45.75 & 313.37 & 59.03 & 50.65 & 335.30 & 198.47 & 72.93 \\
LightGPT (Llama2-7B)     & 275.11 & \underline{161.35} & 46.38  & 269.01 & 102.92 & 43.06 & \underline{\underline{260.53}} & \underline{\underline{127.75}} & 41.84 & 314.24 & 59.59 & 39.66 & 333.94 & 191.63 & \underline{65.49} \\
LightGPT (Llama3-8B)     & \underline{275.10} & 161.92 & 48.25  & 268.81 & 102.54 & 42.74 & 262.29 & 130.32 & 44.96 & \underline{311.72} & 58.40 & 47.60 & 333.85 & 192.06 & 69.75 \\
LightGPT (Llama2-13B)    & \textbf{274.03} & \textbf{159.39} & \textbf{43.24} & \textbf{266.94} & \textbf{100.46} & \underline{40.34} & \textbf{260.17} & \textbf{127.08} & \underline{41.00} & \underline{\underline{310.78}} & \underline{\underline{56.93}} & \underline{\underline{38.64}} & \underline{\underline{330.71}} & \underline{\underline{189.09}} & \underline{\underline{64.16}} \\
\bottomrule
\end{tabular}}
% \vspace{-5pt}
\end{table*}

% \vspace{-5pt}
\subsection{Comparison Between the Traditional Methods and LLMLight (RQ1)}
We first conduct experiments on Jinan and Hangzhou with five real-world traffic flow datasets. The comparison between traditional approaches and LLMLight is shown in Table \ref{tab:overall}. LightGPT refers to the backbone LLMs trained through our proposed fine-tuning method. The experimental results reveal that LLMLight, equipped with LightGPT, consistently achieves state-of-the-art (SOTA) or comparable performance against all baselines.

In detail, most transportation methods lag behind RL and LLMLight. Among RL models, the CoLight series stands out by integrating neighboring information beyond the target intersection. Although Advanced-CoLight, the SOTA RL method, slightly outperforms LLMLight on Hangzhou datasets, it requires communication among adjacent intersections. In contrast, LLMLight achieves competitive results using observation features solely from the target intersection, showcasing its notable effectiveness. To provide a fair comparison, we also compare Advanced-CoLight without the agent cooperation mechanism and LLMLight, detailed in Appendix \ref{subsec:fair_compare}. The results show that LLMLight excels in all metrics under this fair setting. We believe LLMLight can be further improved in the future by incorporating multi-agent cooperation.

In our evaluation of LLMLight powered by generalist LLMs, GPT-4 excels and competes with top RL methods, highlighting the effectiveness of our proposed control framework. Llama3-70B and Qwen2-72B rank second and third. Surprisingly, ChatGPT-3.5 shows the least favorable performance, indicating that generalist LLMs may lack the specialized expertise for optimal traffic management. This suggests a potential need for domain-specific LLMs for intelligent traffic control in the future. Appendix \ref{subsec:failure} provides more performance comparisons and analysis of their limitations.

Examining the models trained by our proposed method (\ie LightGPTs), even LLMLight powered by the smallest model, LightGPT (Qwen2-0.5B), outperforms the much larger Llama2-70B. This highlights the effectiveness of our backbone model training approach. While lightweight models with 7B or 8B parameters exhibit a slight performance drop compared to LightGPT (Llama2-13B), they offer more economical alternatives for deployment. Please refer to Appendix \ref{subsec:cost-effectiveness} for a detailed cost-effectiveness analysis.

\begin{small}
\begin{table}[t]
  \centering
  \caption{Abalation tests of four LightGPT variants.}
  \label{tab:abalation}
  \setlength{\tabcolsep}{3pt}
  \resizebox{\columnwidth}{!}{
    \begin{tabular}{c|cccccc}
    \toprule
    \multirow{3}{*}{Model} & \multicolumn{6}{c}{Average Travel Time (ATT, s)} \cr
    \cmidrule(lr){2-7}
    & GPT-4 & Qwen2-0.5B & Qwen2-7B & Llama2-7B & Llama3-8B & Llama2-13B \cr
    \midrule
    Origin               & 318.71 & 1124.84 & 621.80 & 973.50 & 1136.17 & 722.73 \cr
    IFT (W/O filtering)    & - & \underline{788.93}  & \underline{314.33} & \underline{316.43}  & \underline{316.70} & \underline{313.32} \cr
    IFT                   & - & \underline{\underline{331.32}}  & \underline{\underline{314.32}} & \underline{\underline{315.94}}  & \underline{\underline{314.56}} & \underline{\underline{312.43}} \cr
    IFT + CGPR            & - & \textbf{328.11}  & \textbf{313.37} & \textbf{314.24}  & \textbf{311.72} & \textbf{310.78} \cr
    \bottomrule
    \end{tabular}}
\vspace{-10pt}
\end{table}
\end{small}

% \begin{figure}[t]
% \centering
% \includegraphics[width=\columnwidth]{figs/abalation.png}
% \centering
% \caption{Abalation study.}
% \label{fig:ablation}
% \vspace{-10pt}
% \end{figure}

% \vspace{-5pt}
\subsection{Abalation Study}
To verify the effectiveness of each stage in our proposed backbone LLM training method, we conduct ablation studies on four variants of our proposed LightGPT: (1) Origin: The original LLM without any fine-tuning. (2) IFT (W/O filtering): The LLM trained only with imitation fine-tuning, excluding the high-quality reasoning trajectory filtering strategy. (3) IFT: The LLM trained only with imitation fine-tuning (IFT). (4) IFT + CGPR: The LLM trained through all stages, including imitation fine-tuning (IFT) and critic-guided policy refinement (CGPR). The experiment results on five backbone LLMs are reported in Table \ref{tab:abalation}. The results demonstrate that the effectiveness of each backbone LLM significantly improves after applying our proposed fine-tuning strategy. Furthermore, each training stage consistently enhances the control policy of the LLMs, as evidenced by the progressively improved performance with the incorporation of each proposed stage.

By fine-tuning to imitate the reasoning trajectories of GPT-4, LightGPT demonstrates comparable or even better performance than GPT-4. This highlights the innate capability of the backbone model to excel in traffic signal control tasks if not constrained by their absence of traffic management expertise. Furthermore, The integration of policy refinement corrects the reasoning trajectories and decisions of the backbone LLM, resulting in a discernible enhancement. This implementation attains SOTA or competitive performance across all the datasets and significantly surpasses GPT-4, accentuating the promising prospects for advancing intelligent transportation-oriented LLMs.

\begin{figure}[t]
  \begin{subfigure}{\columnwidth}
    \centering
    \includegraphics[width=\linewidth]{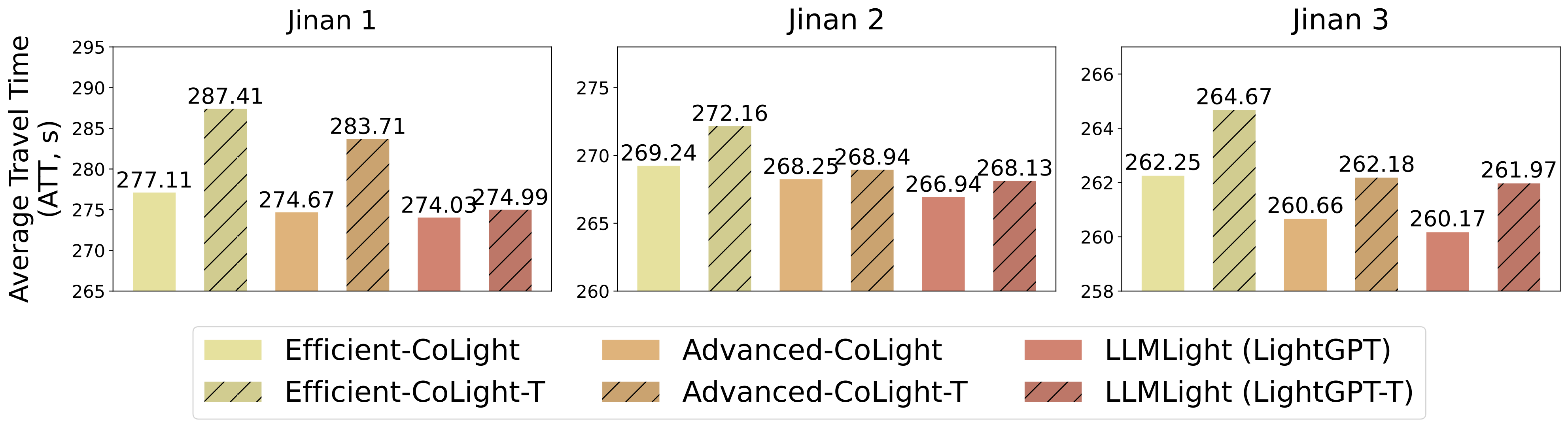}
    \caption{The comparison on Jinan datasets.}
    \label{fig:jinan_trans}
  \end{subfigure}
  \begin{subfigure}{\columnwidth}
    \centering
    \includegraphics[width=\linewidth]{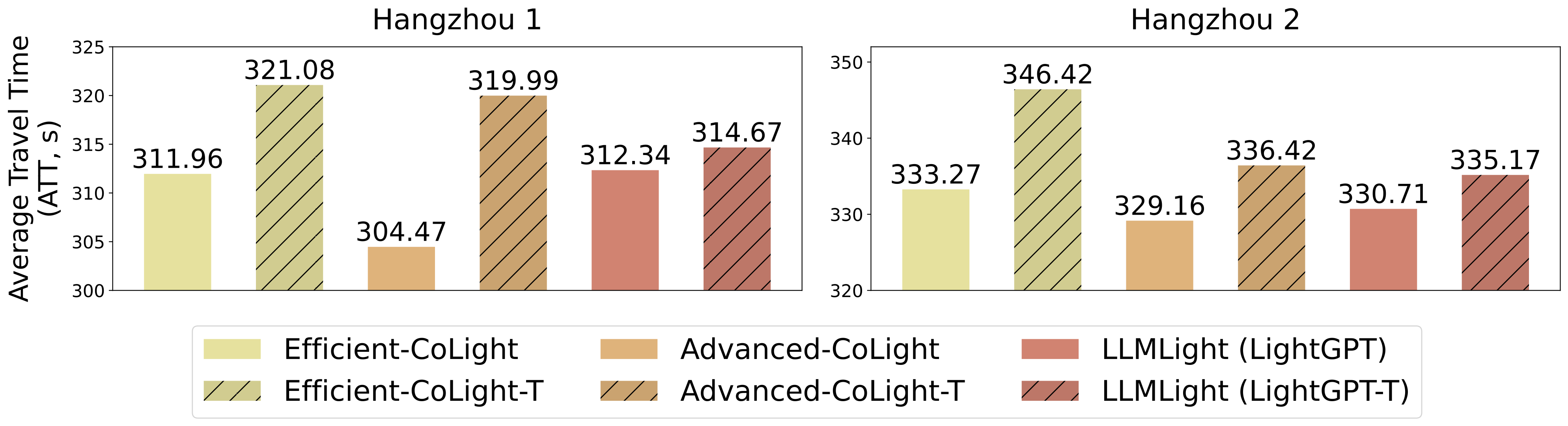}
    \caption{The comparison on Hangzhou datasets}
    \label{fig:hangzhou_trans}
  \end{subfigure}
  \caption{The transferability of different methods.}
  \label{fig:trans}
\vspace{-10pt}
\end{figure}

\begin{figure}[t]
\centering
\includegraphics[width=\columnwidth]{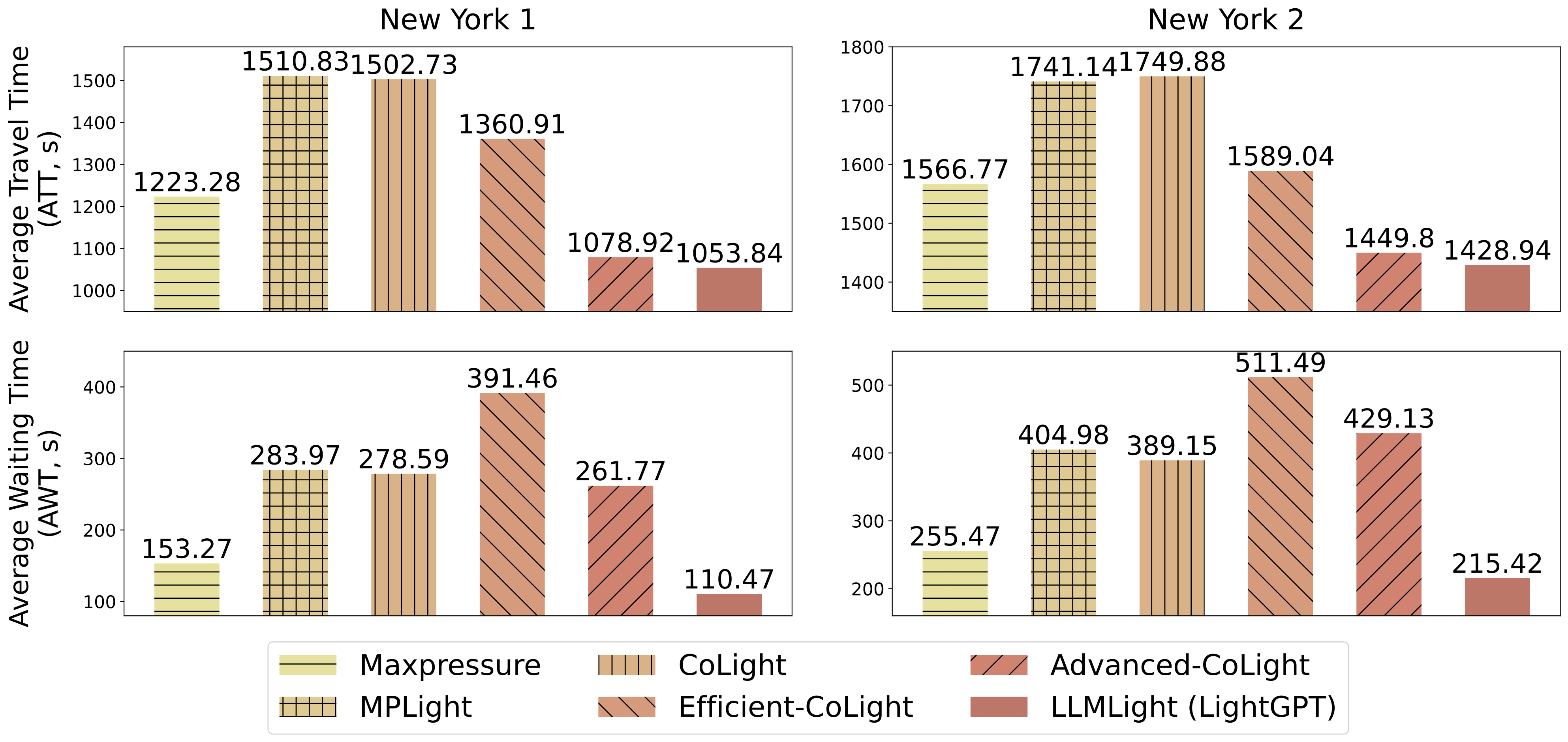}
\centering
\caption{The performance in the large-scale road network.}
\label{fig:newyork}
\vspace{-10pt}
\end{figure}

\subsection{Generlization Comparision (RQ2)}\label{subsec:generalization}
In this section, we evaluate the generalization ability of traditional methods and our proposed LLMLight in unseen environments. LLMLight is powered by the LightGPT pre-trained on Llama2-13B. All models undergo uniform pre-training using either the Jinan 1 or Hangzhou 1 dataset. An additional 24-hour cycle experiment is detailed in Appendix \ref{subsec:24hour}.

\subsubsection{Transferability}
We first study the transferability ability of different methods by implementing the pre-trained model in another distinct road network. The experiment results are shown in Figure \ref{fig:trans}. Models labeled "-T " are pre-trained on a distinct road network (\eg we evaluate the transferability in Hangzhou using the models pre-trained in Jinan). Otherwise, they are trained and tested in the same dataset. We observe a notable decline in the performance of RL-based methods after the transfer, particularly evident in the Jinan 1 and Hangzhou 1 datasets. In contrast, LLMLight consistently achieves superior performance and demonstrates remarkable transferability across all datasets. These findings highlight LightGPT's exceptional generalization capability across diverse traffic contexts and robustness for practical implementations.

\subsubsection{Scalability}\label{subsec:scalability}
Second, we delve into the scalability analysis of both traditional methods and LLMLight, scrutinizing their performance when applied to a considerably larger road network, New York, compared to the original road networks they are pre-trained on (\ie the Jinan 1 dataset). The results of our experiments are detailed in Figure \ref{fig:newyork}. We can observe significant performance degradations happening on most RL methods, as they demonstrate much worse performance than the heuristic method, Maxpressure. While the best RL methods exhibit suboptimal average travel time (ATT) in contrast to LLMLight, it is noteworthy that their policies result in significantly prolonged waiting times. Specifically, the waiting times are 57.80\% and 49.80\% longer than those achieved by ours. This observation suggests that the mechanism of the RL methods focuses on minimizing the overall number of queuing vehicles, possibly at the expense of causing extended waiting times for certain queues with fewer vehicles. The significance of minimizing waiting times cannot be overlooked in practical scenarios, as prolonged waits can induce anxiety and constitute a serious concern about the fairness of different drivers. In contrast, our proposed method ensures both the shortest travel and waiting times, highlighting its remarkable scalability and applicability when transitioning to much larger road networks.

% \begin{small}
% \begin{figure}[t]
% \centering
% \includegraphics[width=0.47\textwidth]{figs/queue_lengths.pdf}
% \centering
% \caption{The queue length distribution.}
% \label{fig:queue_distribution}
% \vspace{-10pt}
% \end{figure}
% \end{small}

\subsubsection{Extreme High-traffic Scenarios}
\begin{figure}[t]
\centering
\includegraphics[width=\columnwidth]{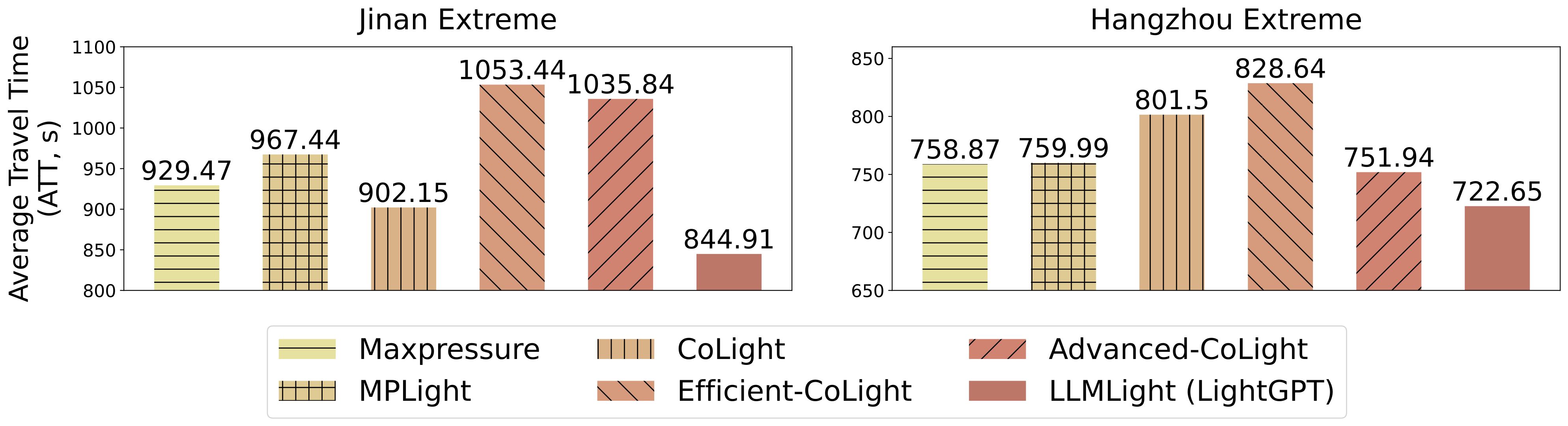}
\centering
\caption{The performance in extreme high-traffic scenarios.}
\label{fig:extreme}
\vspace{-5pt}
\end{figure}

Next, we delve into an uncommon scenario when considerable traffic flow continuously appears at the intersection, a situation that rarely appears during training.
% Figure \ref{fig:queue_distribution} presents the distribution of lane queue lengths across Jinan and Hangzhou datasets, where traffic signals are controlled by Advanced-CoLight. Notably, they demonstrate relatively smooth traffic conditions, with the accumulation of queuing vehicles showing a long-tail distribution.
To assess the efficacy of different methods in extreme high-traffic scenarios, we generate two synthetic traffic flow datasets on the Jinan and Hangzhou road networks, featuring an approximately fourfold increase in vehicles arriving within each 5-minute interval compared to the original datasets. Figure \ref{fig:extreme} shows the performance of both traditional methods and LLMLight armed with LightGPT. All models are pre-trained on regular traffic flow datasets from the same road network (\ie the Jinan 1 or Hangzhou 1 dataset). Similar to experiments on scalability, RL methods also showcase significant performance degradations, which exhibit poorer results than Maxpressure. This suggests that these pre-trained RL methods struggle with out-of-distribution (OOD) traffic conditions when confronted with a significantly increased volume of vehicles compared to their training stage. In contrast, LLMLight consistently shows superior performance, underscoring its robustness and practicality in much heavier traffic conditions.

\begin{figure}[t]
\centering
\includegraphics[width=\columnwidth]{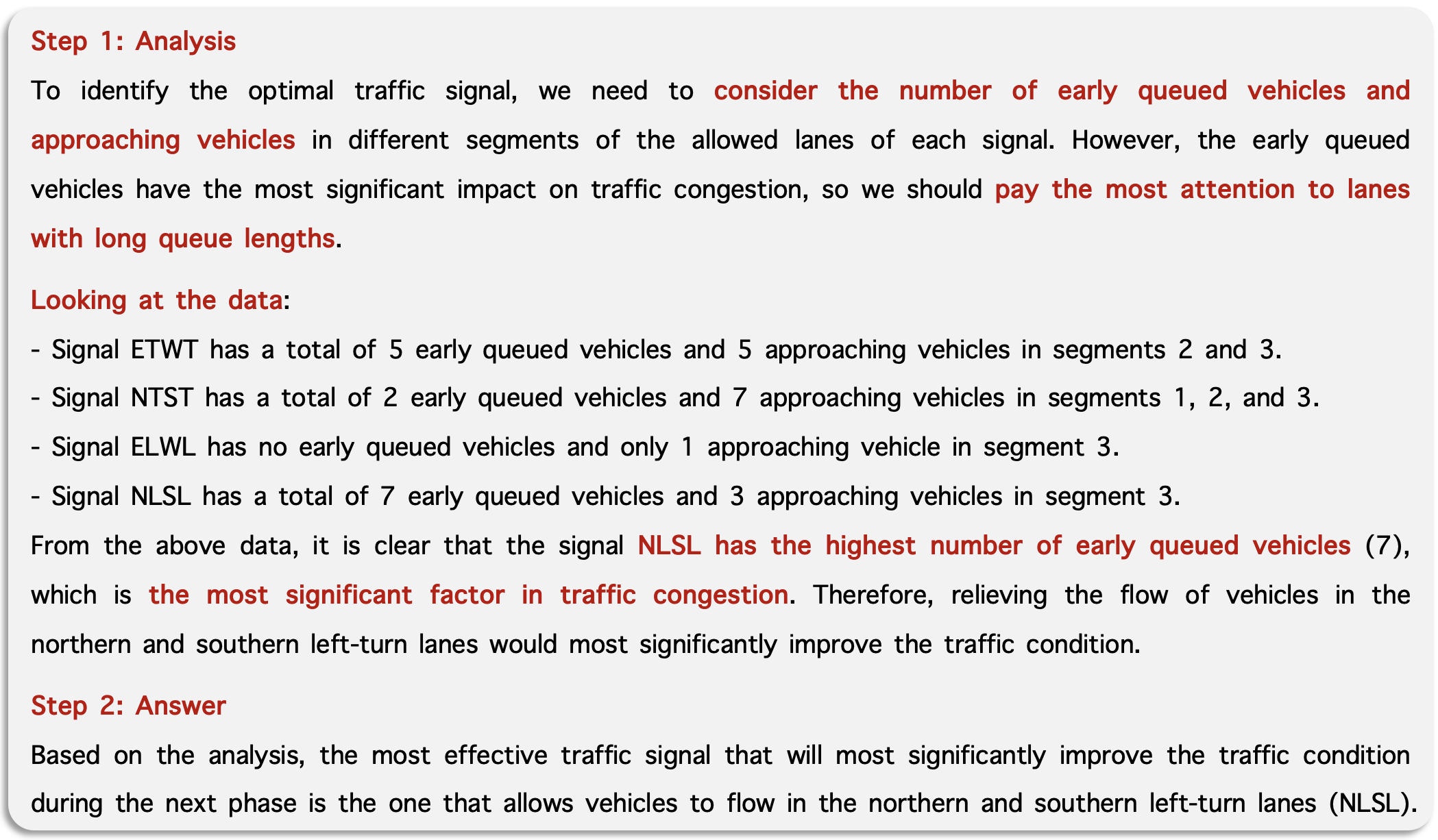}
\centering
\caption{The decision-making process of LightGPT.}
\label{fig:reasoning}
\vspace{-10pt}
\end{figure}

% \vspace{-5pt}
\subsection{Interpretability of LLMLight (RQ3)}\label{subsec:interpret}
% To provide an in-depth assessment of LLMLight's interpretability, we performed a simulation case on the Hangzhou dataset. Figure \ref{fig:reasoning} illustrates the decision-making process of LightGPT (Llama2-13B). Detailed traffic conditions are reported in Appendix \ref{subsec:failure} (Table \ref{tab:case1Traffic}). In this simulated scenario, the north section encounters severe congestion with a significant accumulation of queuing vehicles. LightGPT begins by understanding the task and analyzing the traffic conditions at the targeted intersection. Subsequently, it adeptly identifies that the allowed lanes of signal NLSL demonstrate the heaviest congestion situation, thereby selecting it as the optimal signal phase to set. Unlike RL methods, LLMLight not only excels in making effective control policies but also in providing detailed explanations for each decision. This unique feature enhances the transparency and interpretability of LLMLight, contributing to a more informed understanding of its actions.

To assess LLMLight's interpretability, we analyze a scenario in the Hangzhou dataset. Figure \ref{fig:reasoning} illustrates the decision-making process of LightGPT (Llama2-13B). Detailed traffic conditions are reported in Appendix \ref{subsec:failure} (Table \ref{tab:case1Traffic}). In this scenario, The north section encounters severe congestion with a significant queue. LightGPT first understands the task and analyzes traffic at the intersection. Subsequently, it adeptly identifies that the allowed lanes of signal NLSL demonstrate the heaviest congestion situation, thereby selecting it as the optimal signal phase to set. Unlike RL methods, LLMLight not only excels in making effective control policies but also in providing detailed explanations for each decision. This unique feature enhances the transparency and interpretability of LLMLight, contributing to a more informed understanding of its actions.

% \begin{small}
% \begin{table}[t]
%   \centering
%   \caption{Human expert evaluation results.}
%   \label{tab:abalation}
%   \setlength{\tabcolsep}{3pt}
%   \resizebox{\columnwidth}{!}{
%     \begin{tabular}{c|ccccc}
%     \toprule
%     Choice & Case 1 & Case 2 & Case 3 & Case 4 & Case 5 \cr
%     \midrule
%     Yes, LightGPT has & \multirow{2}{*}{12} & \multirow{2}{*}{12} & \multirow{2}{*}{12} & \multirow{2}{*}{12} & \multirow{2}{*}{12} \cr
%     made the right decision. & & & & & \cr
%     \midrule
%     No, LightGPT does not & \multirow{2}{*}{0} & \multirow{2}{*}{0} & \multirow{2}{*}{0} & \multirow{2}{*}{0} & \multirow{2}{*}{0} \cr
%     make the right decision. & & & & & \cr
%     \midrule
%     Not sure. & 0 & 0 & 0 & 0 & 0 \cr
%     \midrule
%     Success Rate & 100\% & 100\% & 100\% & 100\% & 100\% \cr
%     \bottomrule
%     \end{tabular}}
% \vspace{-10pt}
% \end{table}
% \end{small}

% \vspace{-5pt}
\subsection{Deployment Analysis}\label{subsec: prototype}
%While our current study is grounded in simulated environments, we are actively transitioning toward real-world deployment through collaborations with traffic departments and organizations. We are currently partnering with the Institute of Deep Perception Technology (Jiangsu Industrial Technology Research Institute) in Jiangsu, China, to push online testing and deployment.

We are partnering with the Institute of Deep Perception Technology\footnote{http://www.idpt-jitri.com/} in China to deploy LLMLight in real-world transportation systems\footnote{A prototype system for the public is available at \url{https://lightgpt2024.github.io/LLMLight_Demo/}.}.
% We report the human evaluation results of LightGPT's performance in real-world traffic environments in \TODO{Figure x}.
We conduct a human expert evaluation of LLMLight across five randomly selected traffic scenarios to assess its reasoning and decision-making capabilities in various traffic scenarios. Feedback was gathered from a diverse group: four traffic police officers from the Kunming Southeast Ring Expressway Jinsong Traffic Police Brigade in Yunnan and the Zhongshan Traffic Police Detachment in Guangdong, China, nine experienced drivers, and two AI specialists. The consensus from these evaluations was overwhelmingly positive, with all participants acknowledging LLMLight's reasoning ability and decisions made in all examined scenarios. They highly praised its applicability and notable interpretability in real-world contexts. The detailed evaluation results, studied traffic scenarios, and LightGPT's reasoning processes are shown in Appendix \ref{subsec:human-evaluation}.

Additionally: 
%(1) we have successfully deployed a demonstration system\footnote{Our demo is available at \url{https://lightgpt2024.github.io/LLMLight_Demo/}} to manage traffic signals and observe LightGPT's decision-making. Figure~\ref{fig:deployroad} in Appendix \ref{subsec:demo} shows a snapshot of our demo system. 
(1) We conduct a cost-effectiveness analysis comparing LightGPT with GPT-4, detailed in Appendix \ref{subsec:cost-effectiveness}, which includes training and deployment budgets and efficiency evaluations. 
(2) A 24-hour experiment in Appendix \ref{subsec:24hour} validated LightGPT's robustness under daily long-term traffic fluctuations. 
%(4) Appendix \ref{subsec:human-evaluation} presents a human expert evaluation for accessing LightGPT's reasoning and decision-making in various traffic scenarios. 
(3) In Appendix \ref{subsec:other_consider}, we discuss potential challenges and data privacy preservation methods in real-world deployment. In conclusion, LightGPT consistently demonstrates superior cost-effectiveness and applicability for real-world deployment scenarios.

%%%%%%%%% Related Work %%%%%%%%%
\vspace{-5pt}
\section{Related Work}
\subsection{Traffic Signal Control}
% Traffic signal control (TSC) methods can be broadly categorized into transportation \cite{koonce2008traffic, hunt1982scoot} and reinforcement learning (RL)-based approaches \cite{chen2020toward, wu2023transformerlight}. Initially, FixedTime \cite{koonce2008traffic} employs a fixed cycle length and phase split to regulate traffic signals in a predetermined pattern. Maxpressure \cite{varaiya2013max} greedily selects signal phases to maximize road network throughput. PressLight \cite{wei2019presslight} inherits the concept of Maxpressure and combines it with a RL algorithm to optimize the throughput of the intersection. Additionally, some studies propose to design more representative state features to describe traffic conditions. Advanced-MP \cite{zhang2022expression} studies vehicle movement-level state features, designing Advanced-MP features based on traffic demand.

Traffic signal control (TSC) methods fall into two main categories: transportation methods and reinforcement learning (RL)-based approaches. Transportation methods include FixedTime \cite{koonce2008traffic}, which uses a set cycle length and phase split, and Maxpressure \cite{varaiya2013max}, which optimizes signal phases to maximize network throughput. PressLight \cite{wei2019presslight} extends Maxpressure by integrating RL to enhance intersection throughput. Recent studies also focus on improving state feature representation for better traffic condition modeling. For instance, Advanced-MP \cite{zhang2022expression} develops advanced state features based on vehicle movement and traffic demand.

% \vspace{-5pt}
\subsection{Intelligent Transportation}
Intelligent transportation applications primarily emphasize forecasting \cite{fan2023RBDAT, fliupractical2022, ning2023uukg} and control \cite{sun2023hierachical, xu2022novel}. For instance, \cite{lai2023preference} presents a meta-learning-based framework designed to deliver personalized vehicle energy consumption predictions tailored to individual drivers. Additionally, \cite{zhang2022multi} introduces a dynamic charging pricing model for electric vehicles (EVs) using a multi-agent graph convolutional reinforcement learning approach.

% \vspace{-5pt}
\subsection{Large Language Models for Decision Making}
% LLMs recently emerged and exhibited a remarkable capacity for zero-shot reasoning and generalization. Exploiting this proficiency, numerous studies \cite{fu2023drive, cui2023drive, chen2023driving, wen2023dilu, jin2023surrealdriver, schick2023toolformer, sha2023languagempc, wang2023voyager, wei2024openti, Xu2023PenetrativeAM, jie2024citygpt} have increasingly employed LLMs as central planners in diverse control tasks. For instance, \cite{jie2024citybench} proposes CityBench, an interactive simulator-based evaluation platform, as the benchmark for the capability of LLMs in the urban domain perception-understanding and decision-making tasks.
% % GLAM \cite{carta2023grounding} trains the LLM control agents through interactive engagement with the environment. 
% DriveGPT4 \cite{xu2023drivegpt4} presents an end-to-end autonomous driving system capable of interpreting vehicle actions and addressing diverse questions posed by human users. More studies on utilizing LLMs in intelligent transportation can be found at \cite{zhou2023vision, zhang2023towards}.

LLMs recently emerged and exhibited impressive zero-shot reasoning and generalization abilities. Exploiting this proficiency, numerous studies \cite{fu2023drive, cui2023drive, chen2023driving, wen2023dilu, jin2023surrealdriver, schick2023toolformer, sha2023languagempc, wang2023voyager, wei2024openti, Xu2023PenetrativeAM, jie2024citygpt} have increasingly employed LLMs as central planners in various control tasks. For example, \cite{jie2024citybench} introduces CityBench, an interactive simulation-based benchmark for evaluating LLMs' capabilities in urban domain perception and decision-making. DriveGPT4 \cite{xu2023drivegpt4} showcases an end-to-end autonomous driving system that interprets vehicle actions and responds to user queries. Additional studies on LLMs in intelligent transportation are detailed in \cite{zhou2023vision, zhang2023towards}.
%%%%%%%%% Conclusion %%%%%%%%%
% \vspace{-10pt}
\section{Conclusion and Limitations}\label{sec:conclusion}
% This study introduces LLMLight, a novel framework that leverages large language models (LLMs) as traffic signal control agents. By instructing LLMs to conduct a human-like, step-by-step analysis of current traffic conditions, the intelligent agent could judiciously choose the optimal control action, thereby enhancing the overall traffic efficiency at the intersection. Furthermore, we propose a backbone LLM training procedure for optimizing a traffic signal control-oriented LLM, LightGPT. Through extensive experiments conducted on ten traffic flow datasets, the framework with LightGPT shows remarkable improvements compared to nine baseline methods and ten leading LLMs, underlying its exceptional effectiveness and generalization ability across diverse traffic scenarios.

This study introduces LLMLight, a framework employing LLMs as traffic signal control agents. By performing a human-like, step-by-step analysis of traffic conditions, LLMLight adjusts traffic lights to improve traffic efficiency. We further propose a backbone LLM training procedure for developing the TSC-oriented LLM, LightGPT. Extensive experiments on ten traffic datasets show that LLMLight with LightGPT significantly outperforms nine baseline methods and ten leading LLMs, demonstrating its superior effectiveness, generalization, and interoperability across diverse traffic scenarios.

% Our findings are summarized as follows: (1) Traditional traffic signal control methods often exhibit limited generalization capabilities, leading to suboptimal transferability, scalability, and effectiveness in unconventional scenarios. (2) While general LLMs often lack a deep understanding of traffic management, employing a tailored prompting workflow and domain-specific fine-tuning proves effective in overcoming these limitations and improving the generated control policies. (3) The LLM-empowered traffic signal control not only demonstrates superior performance across diverse traffic scenarios but also can offer detailed explanations for each decision. This would significantly contribute to a more comprehensible and accountable traffic control system.

Our findings include: (1) Traditional traffic signal control methods have limited generalization capabilities, leading to suboptimal performance in unseen traffic scenarios. (2) Generalist LLMs lack in-depth traffic management understanding, but tailored prompting and domain-specific fine-tuning can significantly improve their control policies. (3) LLM-empowered traffic signal control demonstrates superior performance across diverse scenarios and can provide detailed explanations for each decision, contributing to a more comprehensible and accountable traffic control system.

Our limitations include: (1) Our approach focuses solely on traffic features at the target intersection, neglecting multi-agent cooperation mechanisms for global traffic improvement. (2) We do not incorporate additional modality features, such as camera images, to aid the LLM agent in decision-making. (3) The framework primarily regulates vehicle traffic flow and does not consider the impact of other dynamic factors at the intersection, such as pedestrians and bicycles. We plan to address these limitations to enhance the effectiveness of the LLMLight framework in the future.

% \clearpage
\bibliographystyle{ACM-Reference-Format}
\balance
\bibliography{sample-base}

\clearpage
\nobalance
\clearpage
\appendix

\section{Appendix}

\subsection{Settings of Intersection, Lanes, and Signal Phases}\label{subsec:signal_phases}
We present the most used setting of intersection, lanes, and signal phases in Figure \ref{fig:signal_phases}.

\subsection{Compared Models}\label{subsec:baseline}

\subsubsection{Transportation Methods}

\begin{itemize}[leftmargin=0.5cm]
    \item \textbf{Random}: A baseline policy randomly switches signal phases with a fixed duration.

    \item \textbf{FixedTime} \citep{koonce2008traffic}: A policy gives a pre-defined fixed cycle length with phase time, widely implemented in most traffic scenarios.

    \item \textbf{Maxpressure} \citep{varaiya2013max}: The state-of-the-art transportation-based method, which greedily relieves vehicles on the lanes with the highest pressure (a variable derived from the difference between upstream and downstream queue lengths).
\end{itemize}

\subsubsection{RL-based Methods}

\begin{itemize}[leftmargin=0.5cm]
    \item \textbf{MPLight} \citep{chen2020toward}: A reinforcement learning (RL)-based method that utilizes pressure as both observation and reward. It employs FRAP, a network structure specifically designed to manage unbalanced traffic flow, as the foundational model.

    \item \textbf{AttendLight} \citep{oroojlooy2020attendlight}: Leveraging an attention mechanism, this model constructs observation features and predicts phase transition probability.

    \item \textbf{PressLight} \citep{wei2019presslight}: Leveraging the concept of Maxpressure with DRL, optimizing the pressure of the intersection.
    
    \item \textbf{CoLight} \citep{wei2019colight}: This model employs a graph attention network (GAT) to enable communication among neighboring intersections, enhancing coordination and decision-making.
    
    \item \textbf{Efficient-CoLight} \citep{wu2021efficient}: Building upon the CoLight model, it integrates efficient pressure as an observation, optimizing its decision-making capabilities.
    
    \item \textbf{Advanced-CoLight} \citep{zhang2022expression}: Building upon the CoLight model, it incorporates advanced traffic state features, achieving the SOTA performance in traffic signal control tasks.
\end{itemize}

\subsubsection{Large Language Models}

\begin{itemize}[leftmargin=0.5cm]
    \item \textbf{Llama} \citep{llama2}: This is a series of pre-trained and finely-tuned large language models (LLMs) developed by Meta, ranging in scale from 7B to 70B parameters. They are adaptable for various tasks related to natural language generation. We use the second- and third-generation, Llama-2 and Llama-3, in our experiments.

    \item \textbf{Qwen2} \citep{jin2023qwen}: Alibaba's Qwen2 series are LLMs with parameters ranging from 0.5B to 72B, supporting up to 128K tokens context, enhancing capabilities in natural language understanding, code writing, and multi-language processing.
    
    \item \textbf{ChatGPT-3.5} \citep{gpt35}: Developed by OpenAI, ChatGPT-3.5 is a chatbot explicitly designed for engaging in conversations, answering queries, and assisting with diverse tasks. We adopt \texttt{gpt-3.5-turbo-0613} API from OpenAI.
    
    \item \textbf{GPT-4} \citep{gpt4}: The newest advancement in LLM by OpenAI. It is more reliable and creative compared to GPT-3.5. It excels in handling significantly more nuanced instructions and tasks. We utilize \texttt{gpt-4-0613} API in our experiments.
\end{itemize}

% \begin{small}
% \begin{figure}[H]
% \centering
% \includegraphics[width=\columnwidth]{figs/traffic_condition.pdf}
% \centering
% \caption{The traffic condition of the simulation case.}
% \label{fig:traffic_condition}
% % \vspace{-20pt}
% \end{figure}
% \end{small}

\begin{small}
\begin{figure}[t]
\centering
\includegraphics[width=\columnwidth]{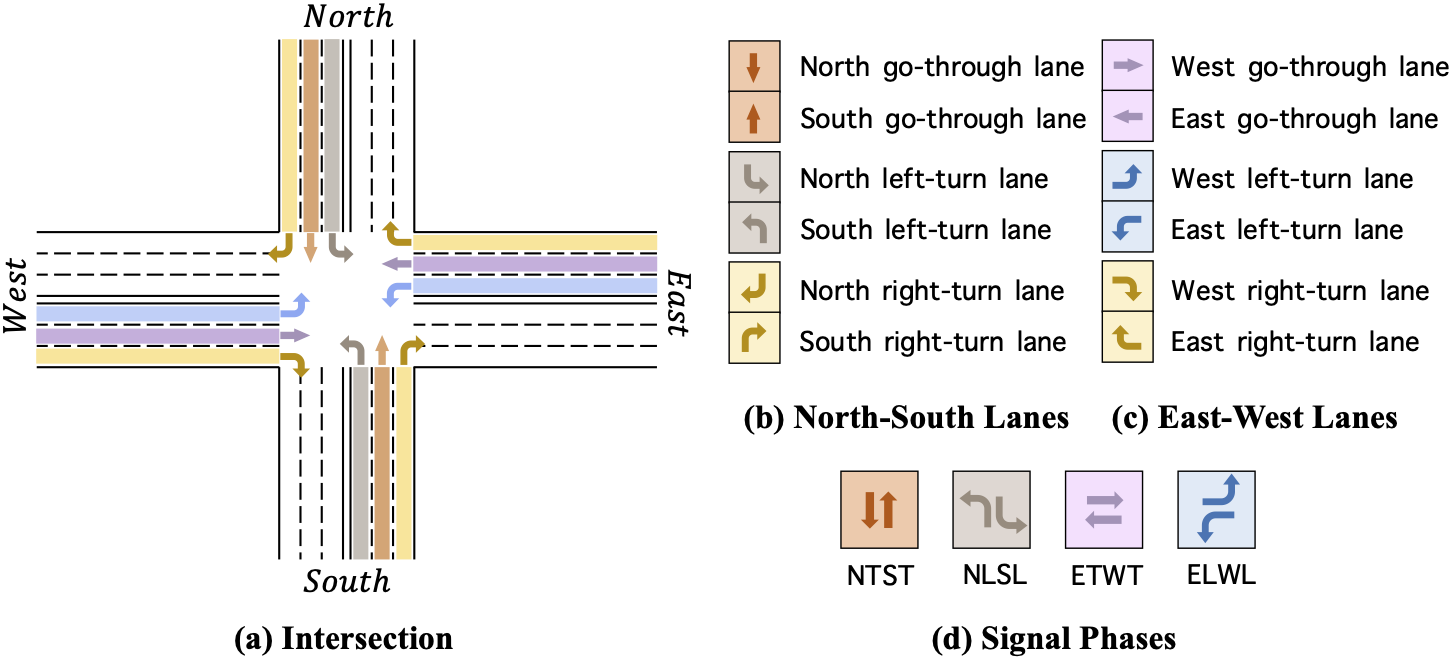}
\centering
\caption{An illustration of intersection, lanes, and signal phases. Lanes sharing a common color signify non-conflicting movements, and the passage is allowed in the corresponding signal phase. Signal phases include ETWT (go-through from east and west), ELWL (left-turn from east and west), NTST (go-through from north and south), and NLSL (left-turn from north and south).}
\label{fig:signal_phases}
% \vspace{-5pt}
\end{figure}
\end{small}

\begin{figure}[t]
\centering
\includegraphics[width=\columnwidth]{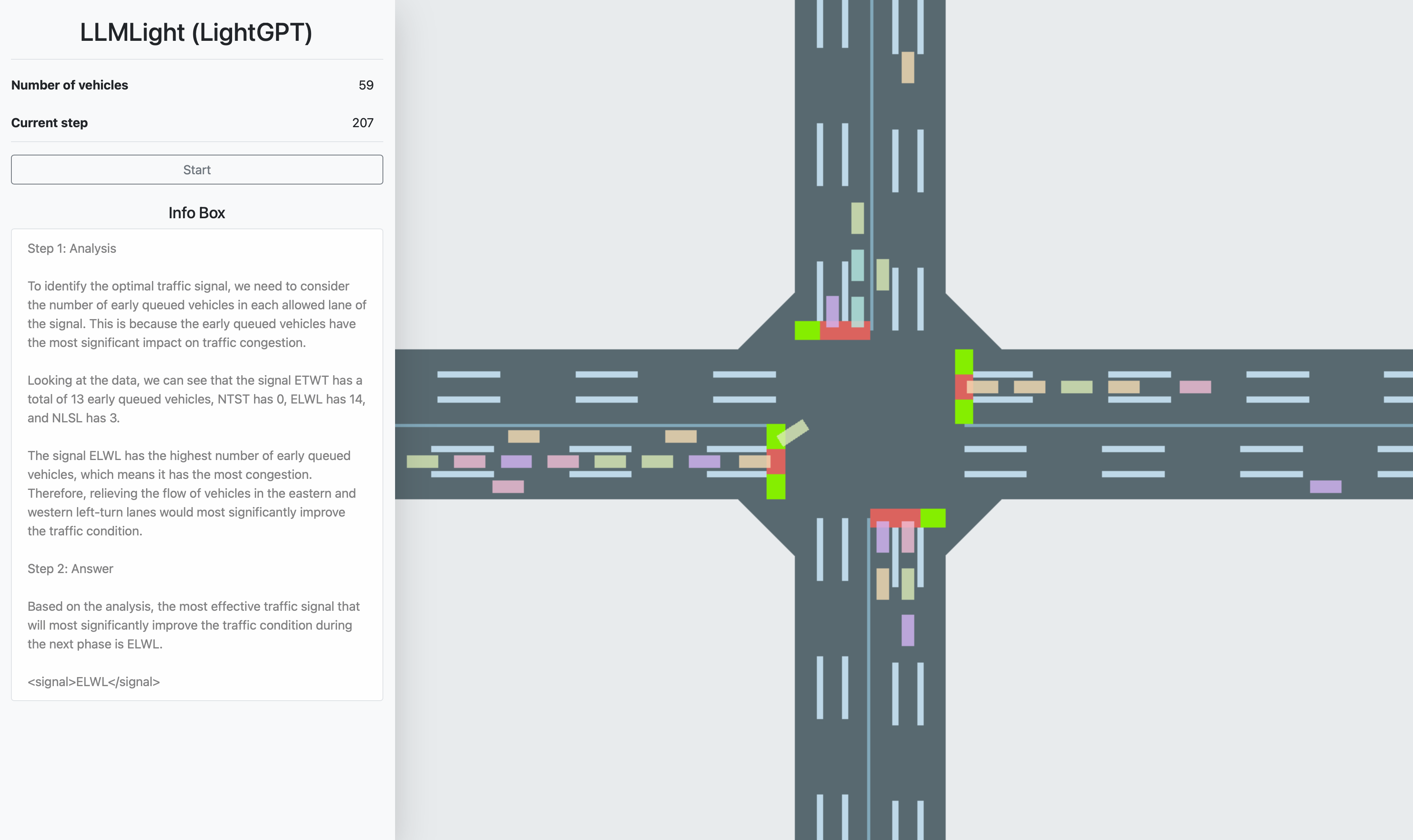}
\centering
\caption{Prototype system.}
\label{fig:deployroad}
% \vspace{-10pt}
\end{figure}

\subsection{Prototype System}\label{subsec:demo}
Figure \ref{fig:deployroad} shows a snapshot of our demo system. We provide users with real-time visualizations of the current traffic conditions on the right, along with statistical data on vehicles in the upper left corner. Additionally, the system further presents the reasoning behind LightGPT's decisions through a message box in the interface's lower left section.

\begin{table*}[t]
\centering
\caption{The annual deployment cost (in USD) of backbone LLMs. LightGPT-8B and LightGPT-13B states LightGPT (Llama3-8B) and LightGPT (Llama2-13B), respectively.}
% \vspace{-5pt}
  \label{tab:deployment-budget}
  \setlength{\tabcolsep}{3pt}
  \resizebox{\textwidth}{!}{
    \begin{tabular}{c|ccc|ccc|ccc|ccc}
    \toprule
    City scale & \multicolumn{3}{c|}{Small} & \multicolumn{3}{c|}{Medium} & \multicolumn{3}{c|}{Large} & \multicolumn{3}{c}{Metropolitan} \\
    \midrule
    \# of traffic lights & \multicolumn{3}{c|}{100} & \multicolumn{3}{c|}{500} & \multicolumn{3}{c|}{1000} & \multicolumn{3}{c}{5000} \\
    \midrule
    Model & GPT-4 & LightGPT-13B & LightGPT-8B & GPT-4 & LightGPT-13B & LightGPT-8B & GPT-4 & LightGPT-13B & LightGPT-8B & GPT-4 & LightGPT-13B & LightGPT-8B \\
    \midrule
    \# of machines & - & 2 & 1 & - & 10 & 5 & - & 20 & 10 & - & 100 & 50 \\
    \midrule
    Annual cost & \$1,680K & \$44.91K & \textbf{\$22.45K} & \$8,400K & \$224.54K & \textbf{\$112.27K} & \$16,800K & \$449.08K & \textbf{\$224.54K} & \$84,100K & \$2,250K & \textbf{\$1,120K} \\
    \bottomrule
    \end{tabular}}
% \vspace{-5pt}
\end{table*}

\begin{small}
\begin{table}[t]
  \centering
  \caption{The training budget (in USD) of LightGPT.}
  \label{tab:training_budget}
  \setlength{\tabcolsep}{3pt}
  \resizebox{\columnwidth}{!}{
    \begin{tabular}{c|cccc}
    \toprule
    & Data Collection	& IFT (1 GPU) & CGPR (4 GPUs) & Sum \cr
    \midrule
    Cost & \$135.53	& \$11.60 & \$5.52	& \$152.65 \\
    \bottomrule
    \end{tabular}}
\vspace{-5pt}
\end{table}
\end{small}

\begin{small}
\begin{table}[t]
  \centering
  \caption{The efficiency (seconds per batch) of backbone LLMs.}
  \label{tab:inference_time}
  \resizebox{\columnwidth}{!}{
    \begin{tabular}{c|ccccc}
    \toprule
    \# of traffic intersections (batch size) & 1 & 5 & 10 & 15 & 20 \cr
    \midrule
    % LightGPT-7B  & 2.80s & 3.75s & 4.20s & 4.65s \cr
    GPT-4        & 8.60s & 9.23s & 11.69s & 11.35s & 12.95s \cr
    LightGPT (Llama2-13B) & 3.53s & 4.61s & 5.31s & 5.74s & 6.62s \cr
    LightGPT (Llama3-8B)  & \textbf{2.52s} & \textbf{3.30s} & \textbf{3.85s} & \textbf{4.15s} & \textbf{4.69s} \cr
    \bottomrule
    \end{tabular}}
\vspace{-10pt}
\end{table}
\end{small}

\subsection{Cost-effectiveness of LightGPT}\label{subsec:cost-effectiveness}
To assess the cost-effectiveness of our proposed LightGPT with the most widely used and advanced LLM, GPT-4. We report the annual deployment cost in different city scales, budget in data collection and model training, and efficiency evaluation of backbone LLMs in Tables \ref{tab:deployment-budget}, \ref{tab:training_budget}, and \ref{tab:inference_time}, respectively. These calculations were performed on a cloud computing machine with 4 × A800-80GB GPUs, 56 cores, 400GB of memory, and a 1TB SSD. Please note that our model weights have been open-sourced. With its exceptional generalization ability, it can be directly deployed in most unseen traffic environments without further fine-tuning. 

For real-time efficiency in deployment, we set the number of parallelly controlled intersections to 20 for LightGPT (Llama3-8B) and 10 for LightGPT (Llama2-13B). When controlling asynchronously, one machine with LightGPT (Llama3-8B) can serve up to 100 intersections, while LightGPT (Llama2-13B) can serve up to 50. We can observe that LightGPT is the most cost-effective solution, as it achieves the best performance in both effectiveness (see Section \ref{sec:experiment}) and real-time traffic control, while also having the lowest deployment cost. Further efficiency and cost-effectiveness improvements are planned for future work.

\subsection{24-hour Cycle Experiment}\label{subsec:24hour}
\begin{figure}[t]
\centering
\includegraphics[width=\columnwidth]{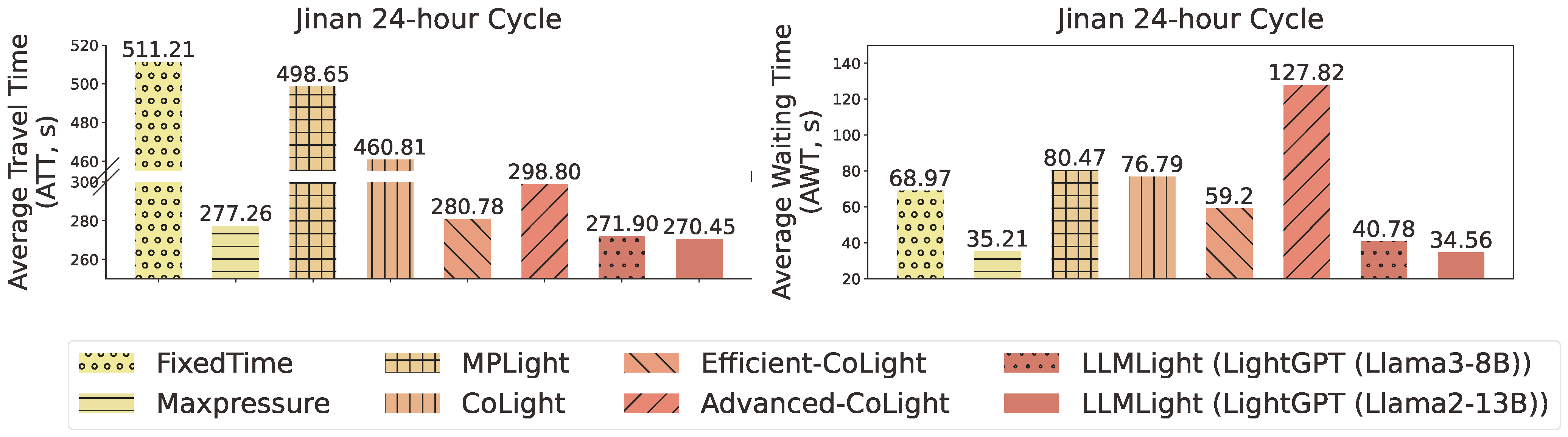}
\centering
\caption{The performance in a 24-hour period.}
\label{fig:24hour}
\vspace{-10pt}
\end{figure}

\begin{figure}[t]
\centering
\includegraphics[width=\columnwidth]{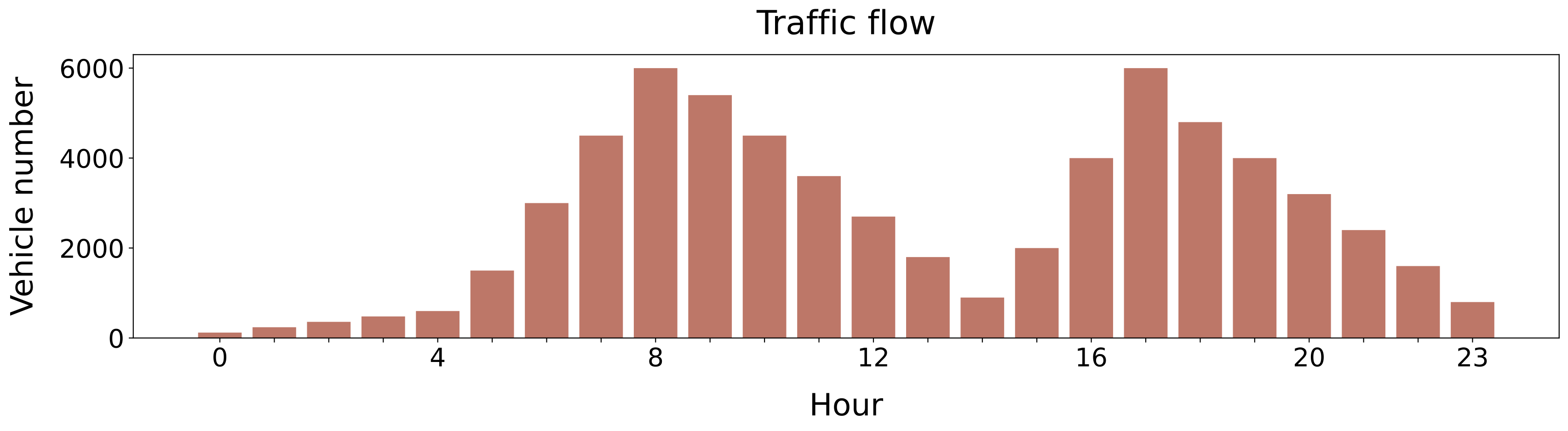}
\centering
\caption{The traffic flow distribution in the 24-hour.}
\label{fig:24hourdistri}
\vspace{-10pt}
\end{figure}

We extend our experiment to a full 24-hour period to examine model performance under daily long-term traffic fluctuations, including distinct morning and evening rush hours. Detailed traffic flow distribution of this dataset can be found in Figure \ref{fig:24hourdistri}. All models are pre-trained using the Jinan 1 dataset. The results are presented in Figure \ref{fig:24hour}. Consistent with previous experiments, we observe that the performance of RL methods drops significantly when traffic flow shifts across different times of the day, as evidenced by their poorer results compared to Maxpressure. As concluded in Subsection \ref{subsec:scalability}, RL methods result in significantly longer waiting times than Maxpressure and LLMLight, due to their focus on minimizing overall queuing vehicles, which may possibly extend the waiting times of queues with fewer vehicles. In contrast, our proposed method consistently ensures both the shortest travel and waiting times, demonstrating its robustness and real-world applicability in diverse and fluctuating traffic conditions. Moreover, LLMLight significantly outperforms the widely used FixedTime method in real-world scenarios, further underscoring the practical benefits of implementing our approach in real-world deployments.

\subsection{Human Expert Evaluation}\label{subsec:human-evaluation}
\begin{figure*}[htbp]
% \vspace{-30pt}
\centering
\includegraphics[width=\textwidth]{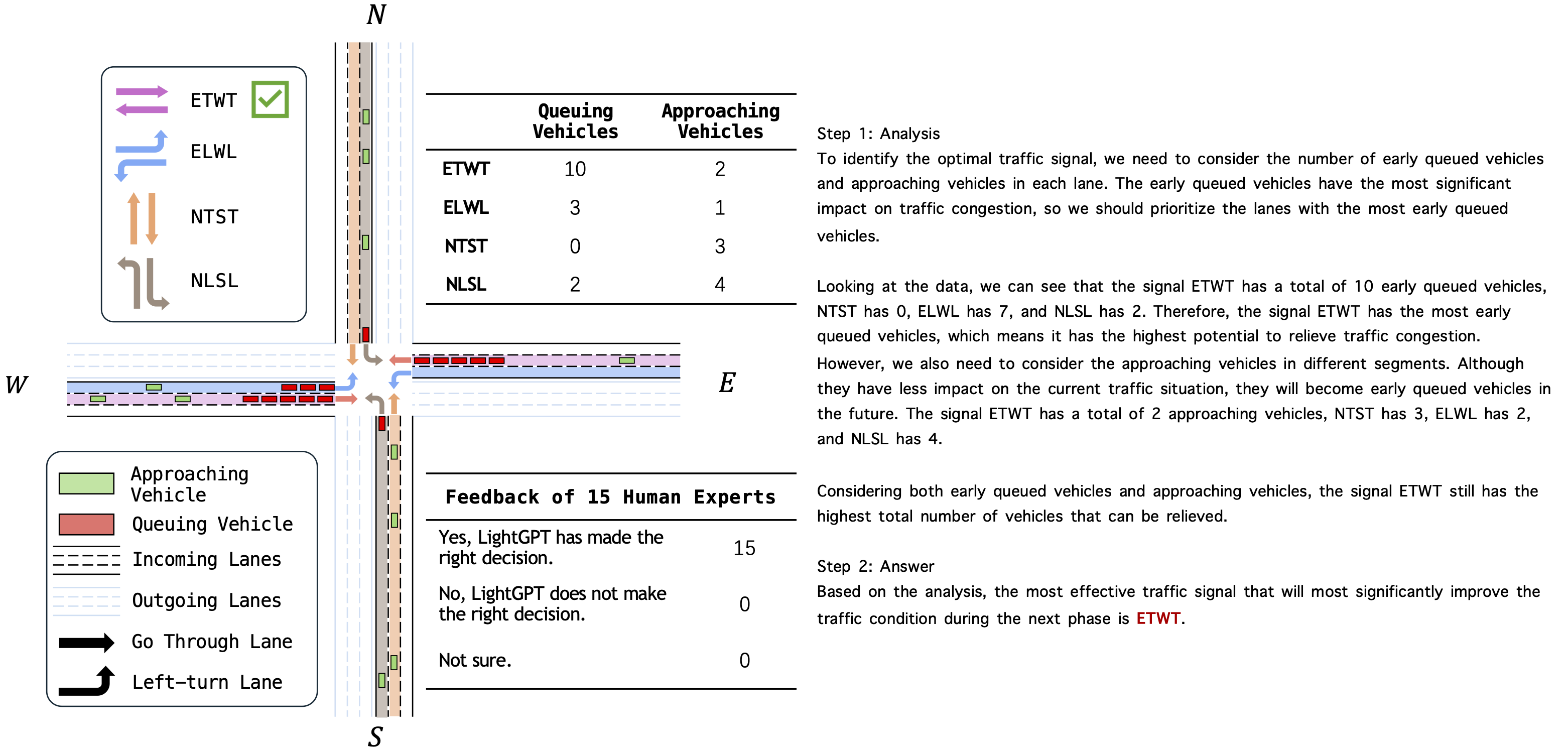}
\centering
\caption{Human expert evaluation: Case 1.}
\label{fig:case1}
% \vspace{-30pt}
\end{figure*}

\begin{figure*}[htbp]
\centering
\includegraphics[width=\textwidth]{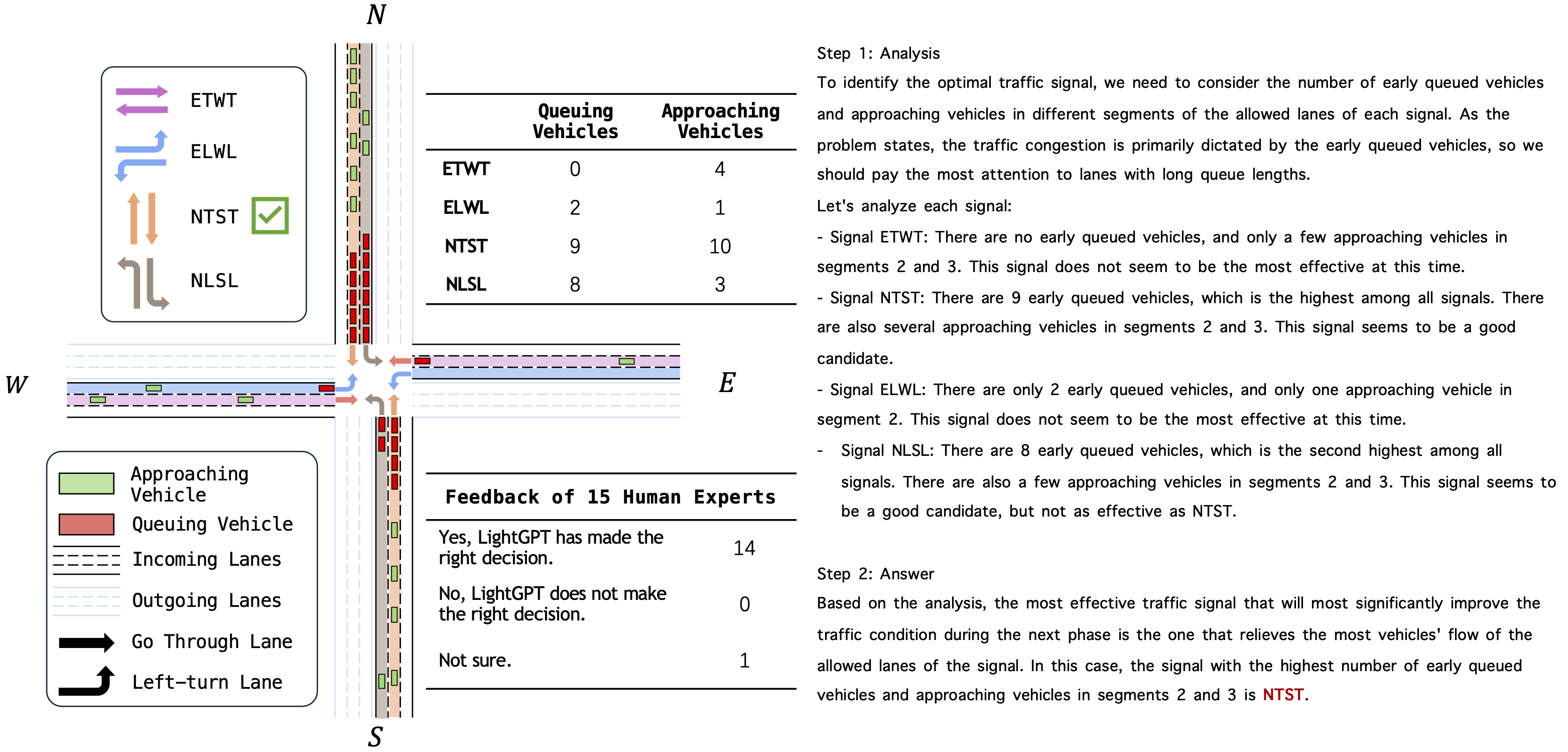}
\centering
\caption{Human expert evaluation: Case 2.}
\label{fig:case2}
% \vspace{-10pt}
\end{figure*}

\begin{figure*}[htbp]
\centering
\includegraphics[width=\textwidth]{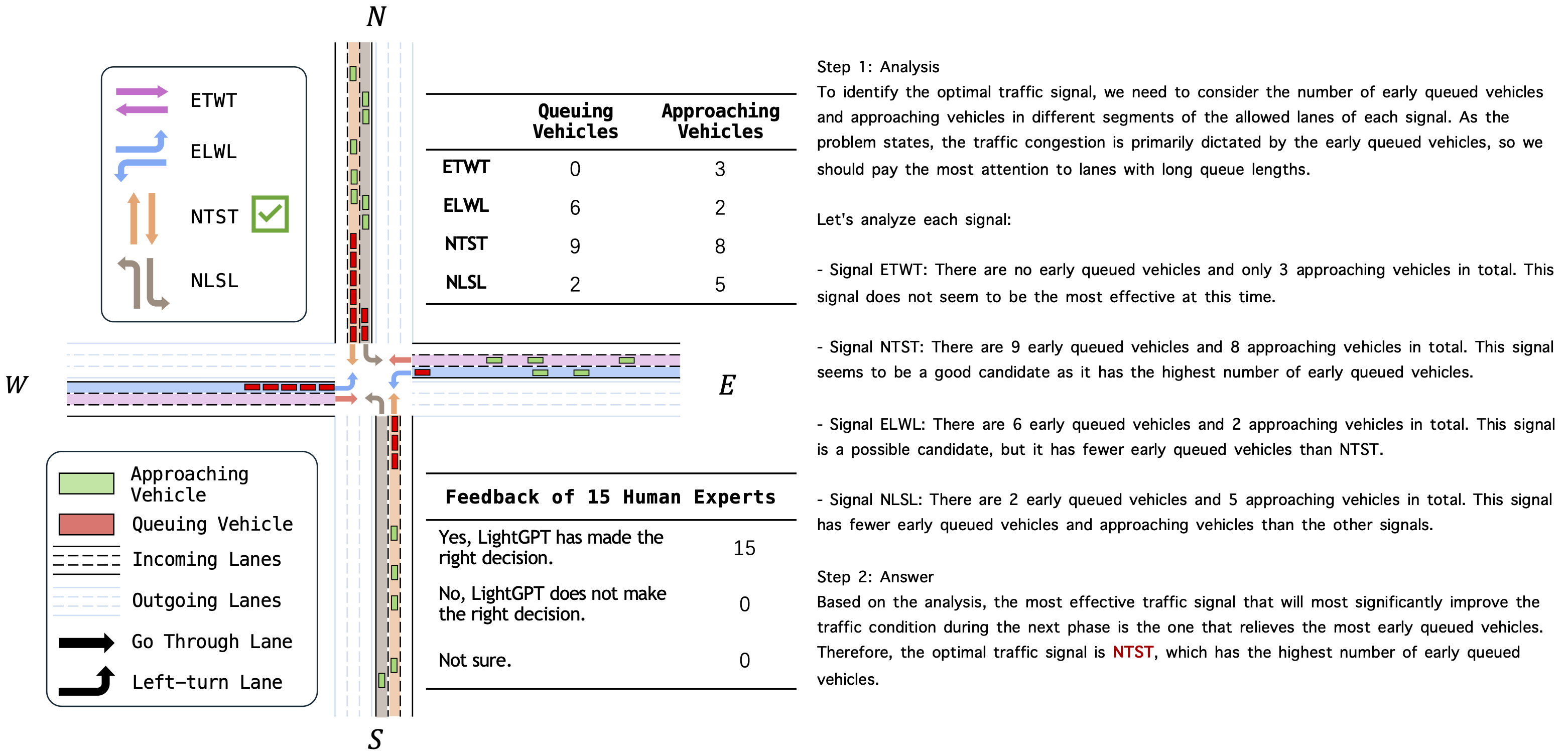}
\centering
\caption{Human expert evaluation: Case 3.}
\label{fig:case3}
% \vspace{-10pt}
\end{figure*}

\begin{figure*}[htbp]
\centering
\includegraphics[width=\textwidth]{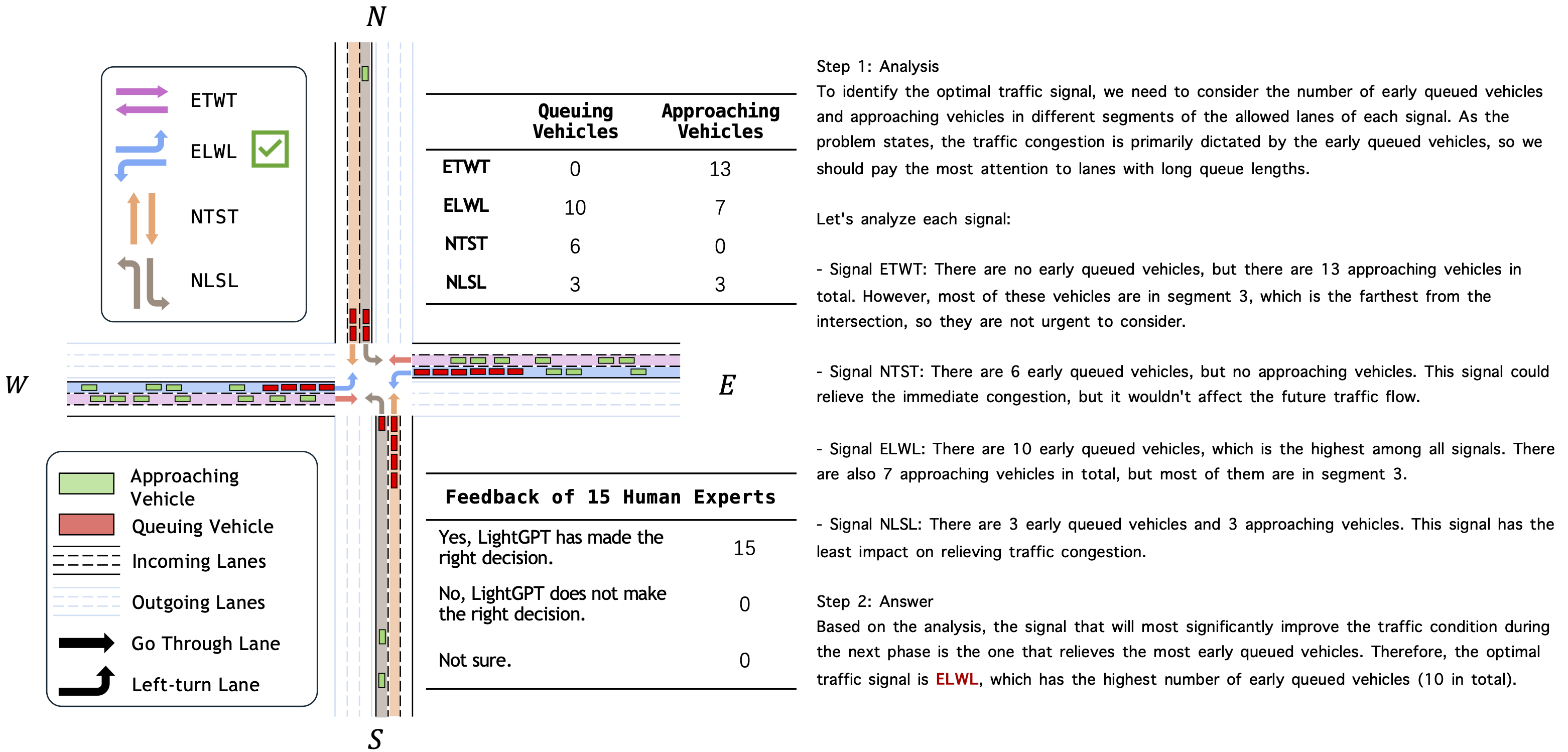}
\centering
\caption{Human expert evaluation: Case 4.}
\label{fig:case4}
% \vspace{-10pt}
\end{figure*}

\begin{figure*}[htbp]
\centering
\includegraphics[width=\textwidth]{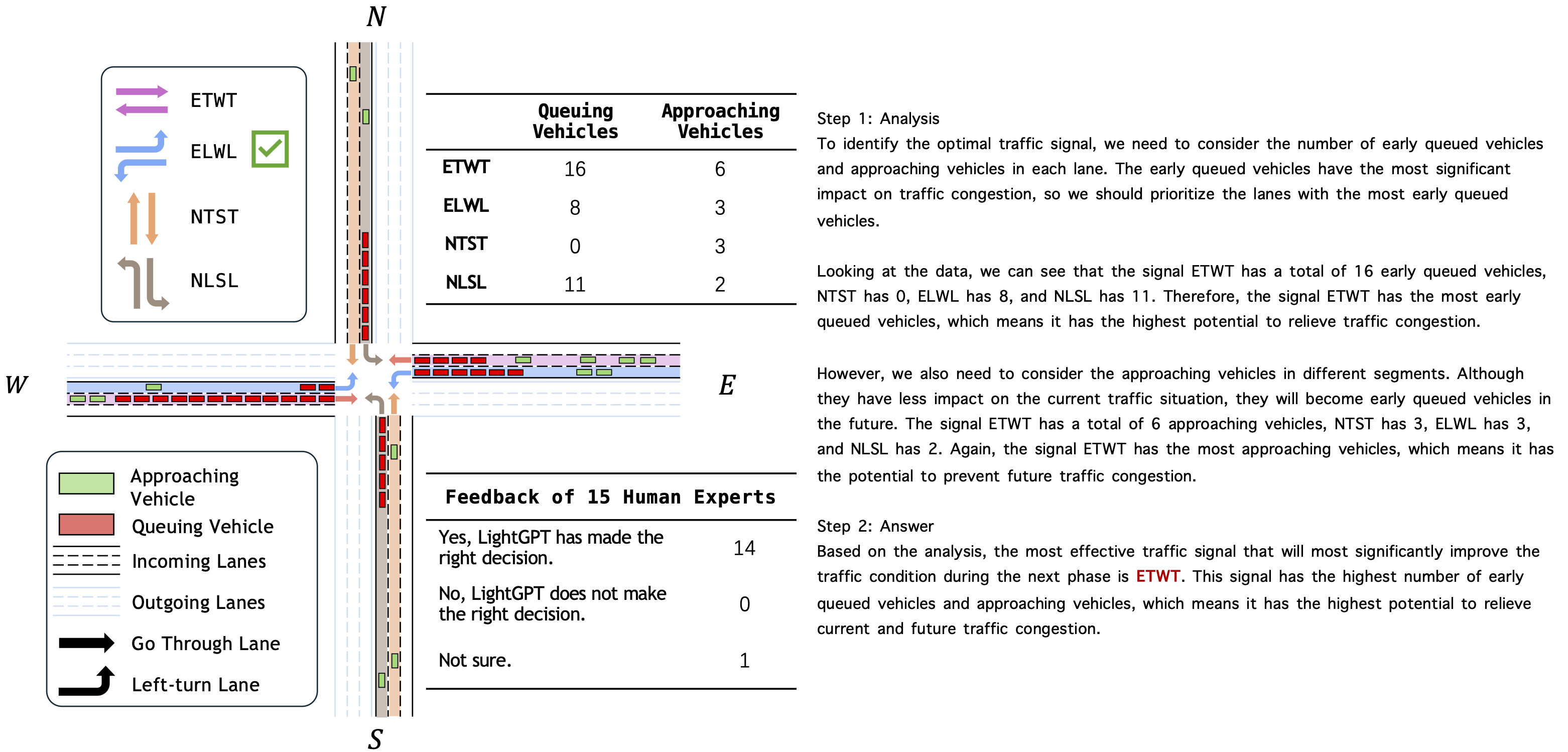}
\centering
\caption{Human expert evaluation: Case 5.}
\label{fig:case5}
\vspace{-5pt}
\end{figure*}

We conduct a human evaluation across five randomly selected traffic scenarios to evaluate LightGPT's reasoning and decision-making capabilities under diverse traffic environments. Feedback was solicited from a diverse group: four traffic police officers from the Kunming Southeast Ring Expressway Jinsong Traffic Police Brigade in Yunnan and the Zhongshan Traffic Police Detachment in Guangdong, China, nine experienced drivers, and two AI specialists. The evaluation results, the studied traffic scenarios, and the reasoning process of LightGPT are presented in Figure \ref{fig:case1}-\ref{fig:case5}. The consensus from these evaluations was overwhelmingly positive, with participants praising its applicability and notable interpretability in real-world contexts.

\subsection{Other Deployment Considerations}

\subsubsection{Challenges in Real-world Deployment}\label{subsec:other_consider}
We outline several challenges that may arise during real-world deployment: (1) \textsl{Integration with existing systems}: LLMLight relies on traffic cameras or ground sensing coils for data collection. While these devices are common in most traffic scenarios, modifications might be needed to integrate our framework seamlessly with current systems. (2) \textsl{Regulatory compliance}: Our experiments have primarily been conducted in China and the U.S., which have different regulatory frameworks compared to other countries. Variations in regulations, such as driver seat positioning and road rules, may require additional adjustments to tailor our model for deployment in various regions. (3) \textsl{Impact on driver behavior}: Unlike traditional traffic signal control (TSC) systems, which follow a predetermined sequence, intelligent TSC systems like LLMLight introduce unpredictability in signal switching. This unpredictability could potentially influence driving behaviors. Further adjustments to address these challenges are planned for future work.

\subsubsection{Data Privacy}
For real-world deployment, we plan to utilize existing infrastructure like traffic cameras and ground sensing coils for observation feature collection. Our system solely extract non-personal data (\ie vehicle counts) to obtain traffic conditions. Any images captured will be securely stored on the traffic department's server and will not be made public, ensuring the maintenance of privacy integrity.

\begin{small}
\begin{table}[t]
  \centering
  \caption{The statistics of the simulation case.}
  % \vspace{-5pt}
  \label{tab:case1Traffic}
  \setlength{\tabcolsep}{3pt}
    \resizebox{\columnwidth}{!}{
    \begin{tabular}{cccc}
    \toprule
    \textbf{Signal} & \textbf{Allowed Lanes} & \textbf{Queue Length} & \textbf{Approaching} \\
    \midrule
    ETWT & East and west go-through & 5 & 5 \\
    \midrule
    ELWL & East and west left-turn & 0 & 1 \\
    \midrule
    NTST & North and south go-through & 2 & 7 \\
    \midrule
    NLSL & North and south left-turn & 7 & 3 \\
    \bottomrule
    \end{tabular}}
% \vspace{-5pt}
\end{table}
\end{small}

\begin{table}[t]
\centering
\caption{The performances of LLMLight with the original LLMs (without fine-tuning) and LightGPTs on the Jinan 1 and Hangzhou 1 datasets.}
% \vspace{-5pt}
\label{tab:other_llm_performance}
\setlength{\tabcolsep}{4pt}
\resizebox{\columnwidth}{!}{
\begin{tabular}{c|ccc|ccc}
\toprule
\multirow{3}{*}{Methods}
& \multicolumn{3}{c}{\textbf{Jinan 1}} & \multicolumn{3}{c}{\textbf{Hangzhou 1}}\cr
\cmidrule(lr){2-4} \cmidrule(lr){5-7}
& ATT & AQL & AWT & ATT & AQL & AWT \\
\midrule
\multicolumn{7}{c}{\textbf{LLMLight (with Generalist LLMs)}} \\
\midrule
Qwen2-0.5B    & 1306.46	& 1771.49 & 953.10 & 1124.84 & 789.44 & 880.79  \\
Qwen2-7B      & 618.86  & 790.53  & 469.08 & 788.98	 & 497.46 & 616.32 \\
Llama2-7B     & 1115.97 & 1580.44 & 896.20 & 973.50  & 664.58 & 819.00 \\
Llama3-8B     & 983.48  & 1236.04 & 830.66 & 1136.17 & 643.44 & 806.26 \\
Llama2-13B    & 508.72  & 567.61  & 169.67 & 722.72  & 418.64 & 311.55 \\
\midrule
\multicolumn{7}{c}{\textbf{LLMLight (with LightGPTs)}} \\
\midrule
LightGPT (Qwen2-0.5B)    & 296.71 & 193.93 & \underline{46.80} & 328.11 & 70.90  & 84.49 \\
LightGPT (Qwen2-7B)      & 275.92 & 163.34 & 48.56 & \underline{313.37} & \underline{59.03}  & 50.65 \\
LightGPT (Llama2-7B)     & \underline{275.11} & \underline{\underline{161.35}} & \underline{\underline{46.38}} & 314.24 & 59.59  & \underline{\underline{39.66}} \\
LightGPT (Llama3-8B)     & \underline{\underline{275.10}} & \underline{161.92} & 48.25  & \underline{\underline{311.72}} & \underline{\underline{58.40}} & \underline{47.60} \\
LightGPT (Llama2-13B)    & \textbf{274.03} & \textbf{159.39} & \textbf{43.24} & \textbf{310.78} & \textbf{56.93} & \textbf{38.64} \\
\bottomrule
\end{tabular}}
% \vspace{-5pt}
\end{table}

\subsection{Failure of Other LLMs}\label{subsec:failure}
\begin{figure*}[!]
\centering
\includegraphics[width=\textwidth]{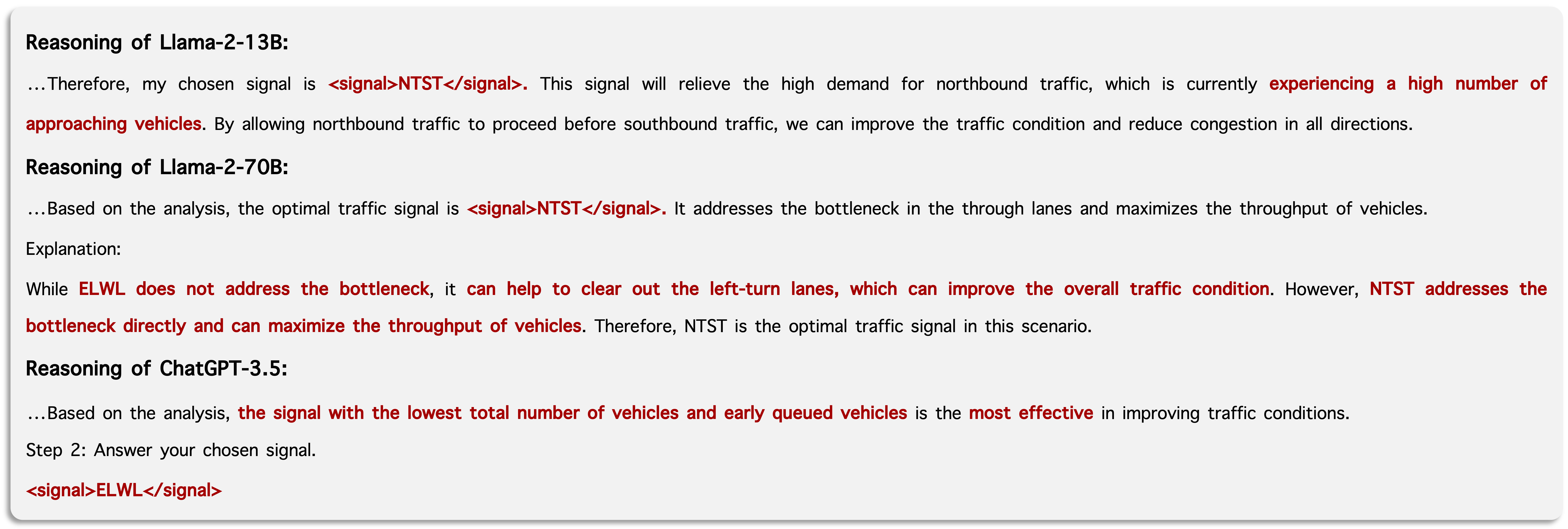}
\centering
\caption{The reasoning of Llama-2 models and ChatGPT-3.5.}
\label{fig:failure}
\vspace{-5pt}
\end{figure*}

To explore the shortcomings of other LLMs, we analyze the reasoning processes of Llama-2 and ChatGPT-3.5, as illustrated in Figure \ref{fig:failure}. Table \ref{tab:case1Traffic} details the traffic conditions, highlighting severe congestion in the north and south sections, with notable queues in the left-turn lane and a heavy influx in the go-through lane. Llama-2 models commonly show limitations in traffic management expertise and instruction-following, often disregarding provided commonsense knowledge. Although they recognize the need to address the most congested lanes, Llama-2 prioritizes approaching vehicles, overlooking the critical queuing vehicles. Conversely, ChatGPT-3.5 exhibits a major logical flaw, claiming that releasing lanes with the lowest vehicle count is most effective for improving traffic conditions. This observation suggests a significant hallucination issue \cite{ji2023survey, huang2023hallucination} in ChatGPT-3.5's traffic signal control reasoning. The performances of more generalist LLMs are detailed in Table \ref{tab:other_llm_performance}.

\subsection{Experiment on Policy Network Learning by Using the Actor-Critic Paradigm}\label{subsec:ppo_exp}
We explore using the actor-critic paradigm, specifically Proximal Policy Optimization (PPO) \cite{schulman2017proximal}, to learn the traffic control policy. Similar to reinforcement learning with human feedback (RLHF) \cite{paul2017deep}, we integrate a multi-layer perceptron (MLP) into the final decoder layer of the LLM as the value function head. The reward function is based on the negative queue length at the target intersection. PPO is employed to facilitate automatic exploration and learning of the optimal control policy and value function. However, this approach struggled to converge and yielded unsatisfactory results, as reported by \cite{wang2023making}. Consequently, we adopted our proposed critic-guided policy refinement (CGPR), which uses the critic model for supervised alignment training, ensuring robust performance. Table \ref{tab:ppo_cgpr_compare} compares the PPO algorithm and our proposed CGPR on the Jinan 1 and Hangzhou 1 datasets.

\begin{small}
\begin{table}[t]
  \centering
  \caption{The comparison between PPO and CGPR.}
  % \vspace{-5pt}
  \label{tab:ppo_cgpr_compare}
  \resizebox{\columnwidth}{!}{
    \begin{tabular}{c|ccc|ccc}
    \toprule
    \multirow{2}{*}{Methods} & \multicolumn{3}{c}{\textbf{Jinan 1}} & \multicolumn{3}{c}{\textbf{Hangzhou 1}} \cr
    \cmidrule(lr){2-4} \cmidrule(lr){5-7}
    & ATT & AQL & AWT & ATT & AQL & AWT \cr
    \midrule
    LLMLight (LightGPT (PPO))	& 275.35 & 164.64 & 57.07 & 315.32 & 60.63 & 50.63 \cr
    LLMLight (LightGPT (CGPR))	& \textbf{274.03} & \textbf{159.39} & \textbf{43.24} & \textbf{310.78} & \textbf{56.93} & \textbf{38.64} \cr
    \bottomrule
    \end{tabular}}
\vspace{-5pt}
\end{table}
\end{small}

\begin{table}[t]
\centering
\caption{The comparison between the SOTA RL and LLMLight without agent cooperation.}
% \vspace{-5pt}
\label{tab:fair_setting}
\setlength{\tabcolsep}{4pt}
\resizebox{\columnwidth}{!}{
\begin{tabular}{c|ccc|ccc}
\toprule
\multirow{3}{*}{Methods}
& \multicolumn{3}{c}{\textbf{Hangzhou 1}} & \multicolumn{3}{c}{\textbf{Hangzhou 2}}\cr
\cmidrule(lr){2-4} \cmidrule(lr){5-7}
& ATT & AQL & AWT & ATT & AQL & AWT \\
\midrule
Advanced-CoLight (W/O COOP)       & \textbf{310.48}	& 58.02	& 54.54	& 332.08	& 192.97	& 88.51 \\
LLMLight (LightGPT (Llama2-13B)) & 310.78 & \textbf{56.93} & \textbf{38.64}	& \textbf{330.71}	& \textbf{189.09}	& \textbf{64.16} \\
\bottomrule
\end{tabular}}
\vspace{-5pt}
\end{table}

% \vspace{-10pt}
\subsection{Comparison with the SOTA RL-based Method Under a Fair Setting}\label{subsec:fair_compare}
Despite Advanced-CoLight's marginal advantage in the Hangzhou datasets, note it requires additional communication among neighboring intersections. However, LLMLight achieves competitive results by leveraging observation features solely from the target intersection. We conducted an ablation study, removing the multi-agent cooperation component from Advanced-CoLight, denoted as Advanced-CoLight (W/O COOP). The experiment results are reported in Table \ref{tab:fair_setting}. In this comparison, LLMLight showcased superior performance in ATT, AQL, and AWT on the Hangzhou dataset. In summary, LLMLight achieves better performance with the same input, and we believe LLMLight can be further improved in the future by incorporating multi-agent cooperation mechanisms.

% \vspace{-10pt}
\subsection{Visualizations of Road Networks}\label{subsec:road_network}
\begin{figure*}[!]
\centering
  \begin{subfigure}{0.32\textwidth}
    \centering
    \includegraphics[width=0.9\textwidth]{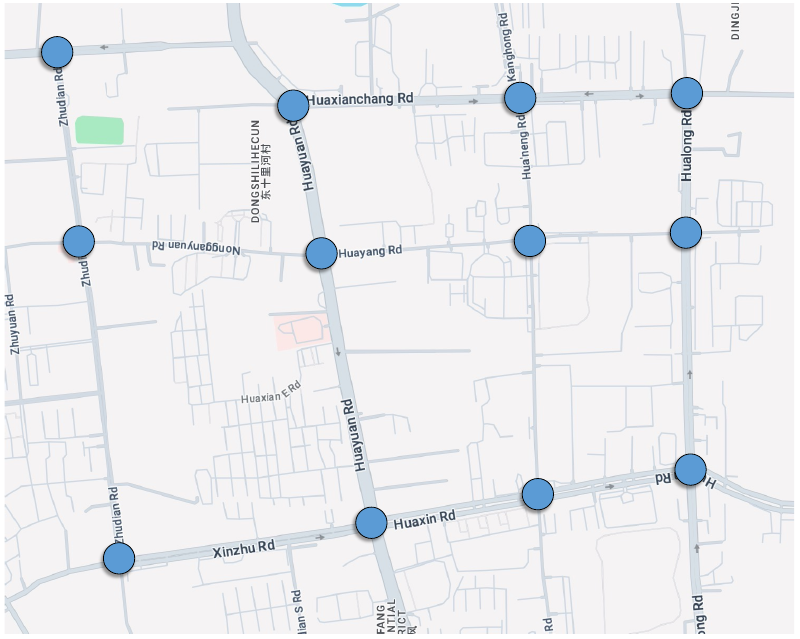}
    \caption{Dongfeng, Jinan, China.}
    \label{subfig:hangzhou}
  \end{subfigure}
  \begin{subfigure}{0.32\textwidth}
    \centering
    \includegraphics[width=0.9\textwidth]{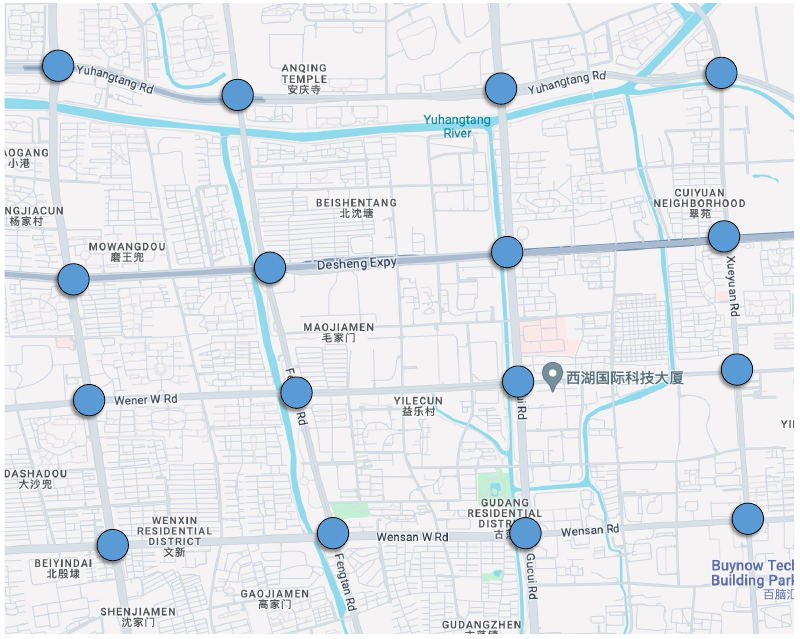}
    \caption{Gudang, Hangzhou, China.}
    \label{subfig:jinan}
  \end{subfigure}
  \begin{subfigure}{0.32\textwidth}
    \centering
    \includegraphics[width=0.9\textwidth]{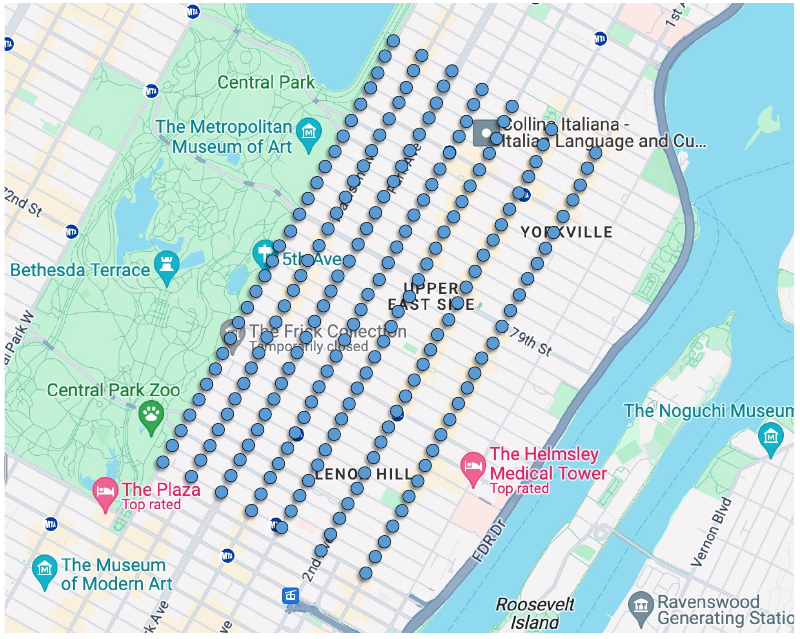}
    \caption{Upper East Side, Manhattan, New York, US.}
    \label{subfig:newyork}
  \end{subfigure}
  \caption{The road network systems of the Jinan, Hangzhou, and New York datasets.}
  \label{fig:roadnet}
% \vspace{-10pt}
\end{figure*}

The visualizations of the road networks are shown in Figure \ref{fig:roadnet}.

% \vspace{-10pt}
\subsection{Prompt Template}\label{subsec:prompt}
We present the prompt template of LLMLight in Table \ref{tab:prompt}.

\afterpage{\clearpage}

\begin{table*}[t]
  \centering
  \caption{The prompt template.}
  \vspace{-5pt}
  \begin{tabularx}{\textwidth}{X}
    \toprule
    % \textbf{Prompt Template} \\
    % \midrule
    \texttt{A traffic light regulates a four-section intersection with northern, southern, eastern, and western sections, each containing two lanes: one for through traffic and one for left-turns. Each lane is further divided into three segments. Segment 1 is the closest to the intersection. Segment 2 is in the middle. Segment 3 is the farthest. In a lane, there may be early queued vehicles and approaching vehicles traveling in different segments. Early queued vehicles have arrived at the intersection and await passage permission. Approaching vehicles will arrive at the intersection in the future.}\\
    \\
    \texttt{The traffic light has 4 signal phases. Each signal relieves vehicles' flow in the group of two specific lanes. The state of the intersection is listed below. It describes:}\\
    \texttt{- The group of lanes relieving vehicles' flow under each traffic light phase.}\\
    \texttt{- The number of early queued vehicles of the allowed lanes of each signal.}\\
    \texttt{- The number of approaching vehicles in different segments of the allowed lanes of each signal.}\\
    \\
    \texttt{Signal: ETWT}\\
    \texttt{Allowed lanes: Eastern and western through lanes}\\
    \texttt{- Early queued: 1 (East), 4 (West), 5 (Total)}\\
    \texttt{- Segment 1: 0 (East), 0 (West), 0 (Total)}\\
    \texttt{- Segment 2: 0 (East), 2 (West), 2 (Total)}\\
    \texttt{- Segment 3: 2 (East), 1 (West), 3 (Total)}\\
    \texttt{Signal: NTST}\\
    \texttt{Allowed lanes: Northern and southern through lanes}\\
    \texttt{- Early queued: 2 (North), 0 (South), 2 (Total)}\\
    \texttt{- Segment 1: 1 (North), 0 (South), 1 (Total)}\\
    \texttt{- Segment 2: 1 (North), 0 (South), 1 (Total)}\\
    \texttt{- Segment 3: 4 (North), 1 (South), 5 (Total)}\\
    \texttt{Signal: ELWL}\\
    \texttt{Allowed lanes: Eastern and western left-turn lanes}\\
    \texttt{- Early queued: 0 (East), 0 (West), 0 (Total)}\\
    \texttt{- Segment 1: 0 (East), 0 (West), 0 (Total)}\\
    \texttt{- Segment 2: 0 (East), 0 (West), 0 (Total)}\\
    \texttt{- Segment 3: 0 (East), 1 (West), 1 (Total)}\\
    \texttt{Signal: NLSL}\\
    \texttt{Allowed lanes: Northern and southern left-turn lanes}\\
    \texttt{- Early queued: 4 (North), 3 (South), 7 (Total)}\\
    \texttt{- Segment 1: 0 (North), 0 (South), 0 (Total)}\\
    \texttt{- Segment 2: 0 (North), 0 (South), 0 (Total)}\\
    \texttt{- Segment 3: 1 (North), 2 (South), 3 (Total)}\\
    \\
    \texttt{Please answer:}\\
    \texttt{Which is the most effective traffic signal that will most significantly improve the traffic condition during the next phase?}\\
    \\
    \texttt{Note:}\\
    \texttt{The traffic congestion is primarily dictated by the early queued vehicles, with the MOST significant impact. You MUST pay the MOST attention to lanes with long queue lengths. It is NOT URGENT to consider vehicles in distant segments since they are unlikely to reach the intersection soon.}\\
    \\
    \texttt{Requirements:}\\
    \texttt{- Let's think step by step.}\\
    \texttt{- You can only choose one of the signals listed above.}\\
    \texttt{- You must follow the following steps to provide your analysis: Step 1: Provide your analysis for identifying the optimal traffic signal. Step 2: Answer your chosen signal.}\\
    \texttt{- Your choice can only be given after finishing the analysis.}\\
    \texttt{- Your choice must be identified by the tag: <signal>YOUR\_CHOICE</signal>.}\\
    \bottomrule
  \end{tabularx}
  \vspace{-10pt}
  \label{tab:prompt}
\end{table*}

\end{document}